\documentclass{article}
\pdfoutput=1 

\usepackage[preprint]{neurips_2026}

\newif\ifchecklist
\checklistfalse

\usepackage[utf8]{inputenc}
\usepackage[T1]{fontenc}
\usepackage{microtype}
\usepackage{hyperref}
\usepackage{url}
\usepackage{booktabs}
\usepackage{amsfonts}
\usepackage{amsmath}
\usepackage{nicefrac}
\usepackage{xcolor}
\usepackage{colortbl}
\usepackage{graphicx}
\usepackage{multirow}
\usepackage{footmisc}
\usepackage{enumitem}
\usepackage{wrapfig}
\usepackage{natbib}
\usepackage{listings}
\usepackage{longtable}
\usepackage{placeins}

\lstdefinestyle{promptstyle}{
  basicstyle=\ttfamily\scriptsize,
  breaklines=true,
  breakatwhitespace=false,
  breakindent=0pt,
  postbreak=\mbox{$\hookrightarrow$\space},
  columns=fullflexible,
  keepspaces=true,
  frame=single,
  framerule=0.4pt,
  framesep=4pt,
  framexleftmargin=2pt,
  framexrightmargin=2pt,
  xleftmargin=4pt,
  xrightmargin=4pt,
  showstringspaces=false,
  aboveskip=10pt,
  belowskip=6pt
}

\newcommand{\bench}{\textsc{MyPCBench}}
\newcommand{\numapps}{17}
\newcommand{\numtasks}{184}
\newcommand{\persona}{Michael Scott}

\title{\bench{}: A Benchmark for Personally Intelligent Computer-Use Agents}

\author{Lawrence Keunho Jang\;
\textbf{Andrew Keunwoo Jang}\;
\textbf{Jing Yu Koh}\;
\textbf{Ruslan Salakhutdinov} \\
Carnegie Mellon University \\
\texttt{\{ljang, rsalakhu\}@cs.cmu.edu}}
\begin{document}

\maketitle

\begin{abstract}
Current benchmarks for computer-use agents evaluate models in impersonal environments. This leaves a gap between evaluation and deployment where personal assistants are expected to work across a user's whole digital life, including their context, historical data, and logged-in accounts. This gap is widest on web tasks, where live web evaluations cannot exercise sites that require logging in or personal information, the kind of site a real personal assistant has to drive. We introduce \bench{}, which tests computer-use agents as personal assistants on a Linux desktop populated with \numapps{} simulated real-world web applications and a full desktop stack, all seeded for one canonical persona, Michael Scott from \textit{The Office}\footnote{\label{fn:office}\textit{The Office} (US), an American mockumentary sitcom developed by Greg Daniels and aired on NBC, 2005--2013. \url{https://www.imdb.com/title/tt0386676/}.}.
We define \numtasks{} tasks in this environment, each inspired by a real request drawn from the OpenClaw community, and benchmark six closed- and open-weight models with a uniform computer+bash tool surface. We find that the best model, Claude Opus 4.6, fully solves 55.4\% of the tasks, the only model above 50\%. Model failures cluster on tasks that span many applications and on long trajectories, where personalization stresses an assistant the most. We release the environment, task set, and agent harness at \url{https://mypcbench.com}.
\end{abstract}

\begin{figure*}[t]
  \vspace{-0.35in}
  \centering
  \includegraphics[width=\textwidth]{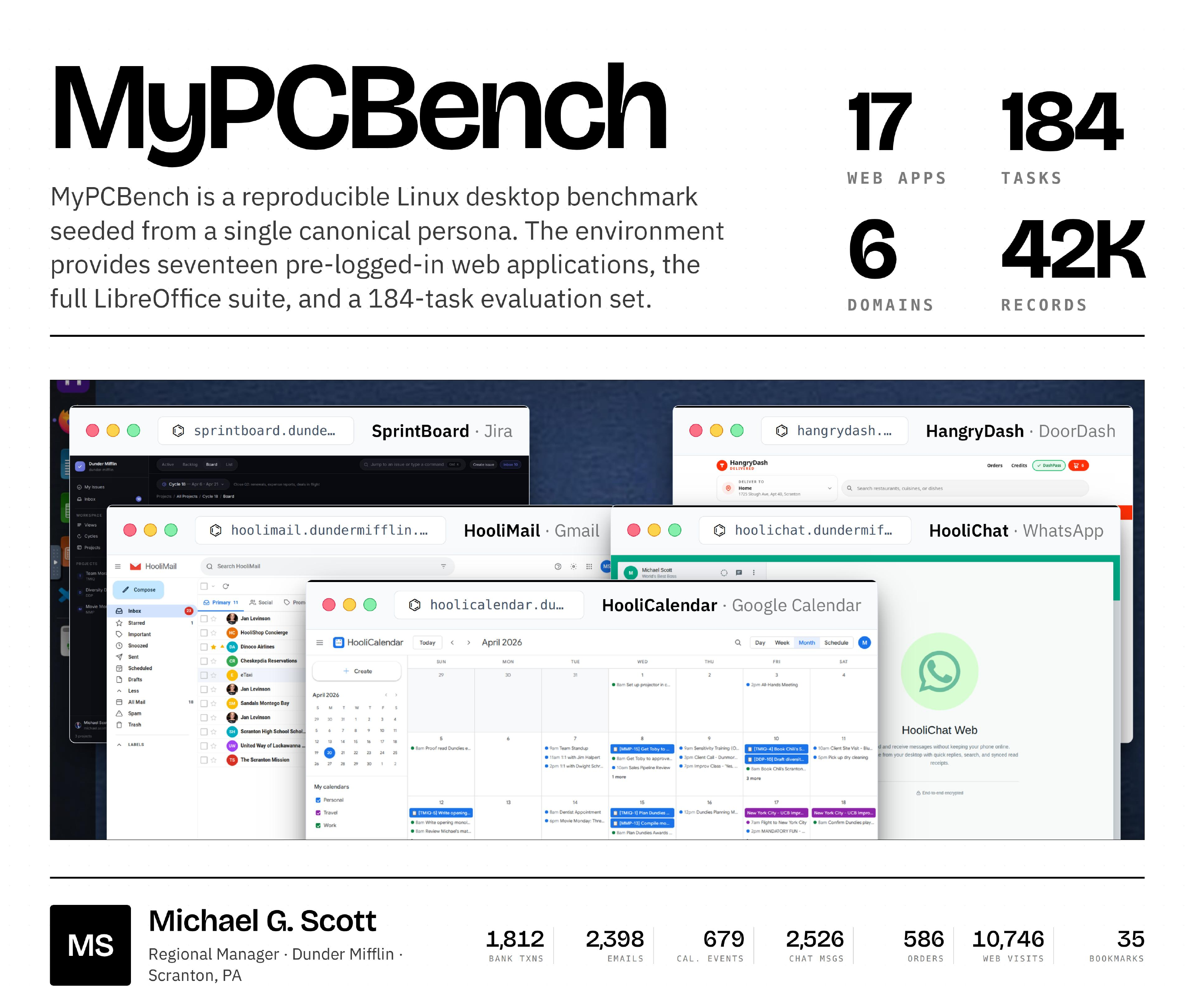}
  \vspace{-0.18in}
  \caption{\small \textbf{Overview of \bench{}.} A reproducible Linux-desktop benchmark for personally intelligent computer-use agents, seeded end-to-end from a single canonical persona (Michael Scott). The image hosts \numapps{} pre-logged-in web apps mirroring real consumer products plus the full LibreOffice suite. The persona's records (itemized in the bottom strip, from 1{,}812 bank transactions to 10{,}746 web visits) are cross-linked so that one trip leaves correlated records across every app that would plausibly book it.}
  \label{fig:hero}
  \vspace{-0.20in}
\end{figure*}

\section{Introduction}
\label{sec:intro}

A person's computer is not a blank slate. Bank transactions, calendar events, email threads, travel bookings, and work chats accumulate across many applications, together making up a user's working and personal record. Current benchmarks for computer-use agents ignore this, with tasks running against empty desktops, generic application states, and minimally seeded databases. In most tasks, the agent is told exactly which application to open and the exact workflow to complete, without realistic user data behind the application. An agent that can place a delivery order but cannot find which restaurant the user actually orders from every Friday has not demonstrated useful capabilities as a personal assistant.
As LLM-based assistants for personal computers move from research demos to consumer products (e.g., OpenClaw~\citep{OpenClaw} and Claude Cowork~\citep{anthropic2026ClaudeCoWork}), evaluation must follow. It should test whether these systems are actually personal and whether personalization performance improves or regresses with each new model release.

Existing agent benchmarks span the web~\citep{zhou2023webarena,koh2024visualwebarena,deng2023mind2web,he2024webvoyager}, full desktops~\citep{xie2024osworld,bonatti2024windows,yang2025macosworld}, enterprise platforms~\citep{drouin2024workarena}, and mobile devices~\citep{rawles2024androidworld}. Most are \emph{simulated} so that grading is deterministic and reproducible. The cost of that reproducibility is impersonality. Each application carries only the data the current task literally needs, and there is no user history behind it. Live-web evaluations in particular avoid sites that require logging in or variable personal information, and the benchmarks that do provide logged-in states (self-hosted WebArena sites, AppWorld's API-level accounts) seed generic or minimal personal history rather than a deep, cross-application personal identity. We argue that this rules out a large fraction of what real users ask their assistants to do.\looseness=-1

No prior benchmark seeds a coherent user identity at the scale of a full personal computer. AppWorld comes closest, seeding accounts at the API layer rather than a lived-in desktop, \S\ref{sec:related}. \bench{} closes this gap. A single persona specification and a deterministic multi-application generator produce an environment that is personal, consistent across applications, and reproducible. Our contribution is that agents must operate over persistent identity, cross-app history, and we provide rubric-graded visible side effects for partial credit under resettable desktop control.

Our canonical persona is \persona{}, the regional manager of a paper company in Scranton, Pennsylvania. Michael's desktop is seeded with 1{,}812 bank transactions, 2{,}398 emails, 679 calendar events with weekly recurrence, 2{,}526 chat and workplace messages, 126 rideshare requests, 402 food-delivery orders, 155 retail orders, 29 grocery orders, and 32 restaurant reservations. A Firefox profile adds 35 bookmarks and 10{,}746 page-history visits, distributed across \numapps{} pre-logged-in web applications and the surrounding desktop stack (Figure~\ref{fig:hero}). The \numapps{} apps and \numtasks{} tasks were chosen through internal author discussion and by manually inspecting the OpenClaw Discord, the largest personalized-LLM-agent community we are aware of, so that \bench{} reflects the types of requests users actually issue to a personal assistant. We make three contributions.

\begin{enumerate}[nosep,leftmargin=*]
  \item A reproducible, cross-app-consistent desktop environment for evaluating personalized agents, built from \numapps{} custom web apps and a full Linux desktop including Firefox, LibreOffice, and file manager, deterministically populated from one persona seed and packaged as a Docker container.
  \item \numtasks{} tasks inspired by real OpenClaw personal-assistant requests, each with a natural-language rubric, plus an agent harness that drives the standard CUA ReAct~\citep{yao2023reactsynergizingreasoningacting} loop against the environment and a rubric-grading LLM-as-a-judge evaluation format.
  \item Benchmarking of six closed- and open-weight models: Claude Opus 4.6 / Sonnet 4.6, GPT-5.5 / GPT-5.4 mini, Qwen 3.5 35B-A3B / 9B under each provider's native computer-use agent with a uniform \texttt{computer}+\texttt{bash} tool surface, with a failure taxonomy and two scaling analyses across trajectory length and number of apps per task.
\end{enumerate}

Our headline finding is that even the strongest current frontier agent (Claude Opus 4.6) fully solves only 55.4\% of \bench{} tasks, and only 36\% of tasks that span 7 or more applications. With every model using the same \texttt{computer}+\texttt{bash} tool action space, GPT-5.5 perfects just 4.5\% of that 7+-app slice and GPT-5.4 mini, Qwen 3.5 35B-A3B, and Qwen 3.5 9B reach 0\%.

\section{Related Work}
\label{sec:related}

\paragraph{Web and desktop agent benchmarks.} The first web benchmarks began with purely synthetic environments (MiniWoB++~\citep{liu2018reinforcement}, WebShop~\citep{yao2022webshop}) and progressed to synthetic realistic websites (WebArena~\citep{zhou2023webarena}, VisualWebArena~\citep{koh2024visualwebarena}) and static datasets of real-website tasks (Mind2Web~\citep{deng2023mind2web}), and then to live Internet evaluations (WebVoyager~\citep{he2024webvoyager}, Online-Mind2Web~\citep{xue2025illusion}). Desktop benchmarks such as OSWorld~\citep{xie2024osworld} extend evaluation to Linux desktops with manually handcrafted reward verifiers. Windows Agent Arena~\citep{bonatti2024windows} and MacOSWorld~\citep{yang2025macosworld} cover the other major operating systems. OpenClaw-inspired benchmarks, including Claw-Eval~\citep{ye2026clawevaltrustworthyevaluationautonomous}, ClawBench~\citep{zhang2026clawbenchaiagentscomplete}, and WildClawBench~\citep{ding2026wildclawbench}, extend previous desktop agent benchmarks by evaluating models on long-horizon, multi-turn, OpenClaw-style use cases. Static grounding benchmarks such as ScreenSpot-Pro~\citep{li2025screenspotproguigroundingprofessional} and OmniACT~\citep{kapoor2024omniactdatasetbenchmarkenabling} evaluate an LLM's ability to ground actions on desktop, browser, and mobile interfaces. A small body of prior work evaluates LLM agents with assigned personas, but it stays inside enterprise contexts. TheAgentCompany~\citep{xu2025theagentcompanybenchmarkingllmagents} places agents in the role of an employee at a simulated software company, and WorkArena~\citep{drouin2024workarena} drives ServiceNow workflows. Generative Agents~\citep{park2023generative} studies a society of agents in a text-based environment and how they interact under separate personas. GAIA~\citep{mialon2024gaia} probes general-assistant competence with no user identity at all. AppWorld~\citep{trivedi2024appworldcontrollableworldapps} seeds a coherent user identity in the API layer, populating 9 simulated apps with one supervisor and a contact network for code agents.\looseness=-1

\paragraph{Personalization benchmarks.} LaMP~\citep{salemi2024lamp} scores LLMs on classification and generation over retrieved user profiles and LongMemEval~\citep{wu2025longmemeval} tests whether chat assistants recall facts from long conversational histories. On the agent side, PersonalWAB~\citep{cai2025personalwab} attaches user profiles and behavior logs to web agents at the web-function layer, and Persona2Web~\citep{kim2026persona2web} grounds web tasks in long-span user histories. All of these hand the model its personal context as an explicit profile, memory store, or interaction history, whereas \bench{} seeds the identity into a full desktop, computer-use evaluation where the personal data lives inside the entirety of the environment.\looseness=-1

\paragraph{Addressing the personalization gap.} Most of the benchmarks above run in the impersonal regime described in \S\ref{sec:intro}, excluding tasks that require personal data or pages behind a login (calling an Uber, paying back a friend on Zelle, reordering the usual DoorDash). \bench{} keeps the OSWorld recipe of a fixed VM image with deterministic snapshot reset, but seeds the desktop end-to-end with Michael Scott's data across every application rather than only the data each task touches. \bench{} pins a single user identity and spans the consumer applications a personal computer actually runs (banking, travel, food delivery, calendar, messaging) on a desktop, making it a benchmark for personal-assistant computer use rather than a stock-state desktop test.\looseness=-1
\section{\bench{}}
\label{sec:env}

\subsection{Environment}
\label{sec:design}

\begin{figure}[t]
  \centering
  \includegraphics[width=\textwidth]{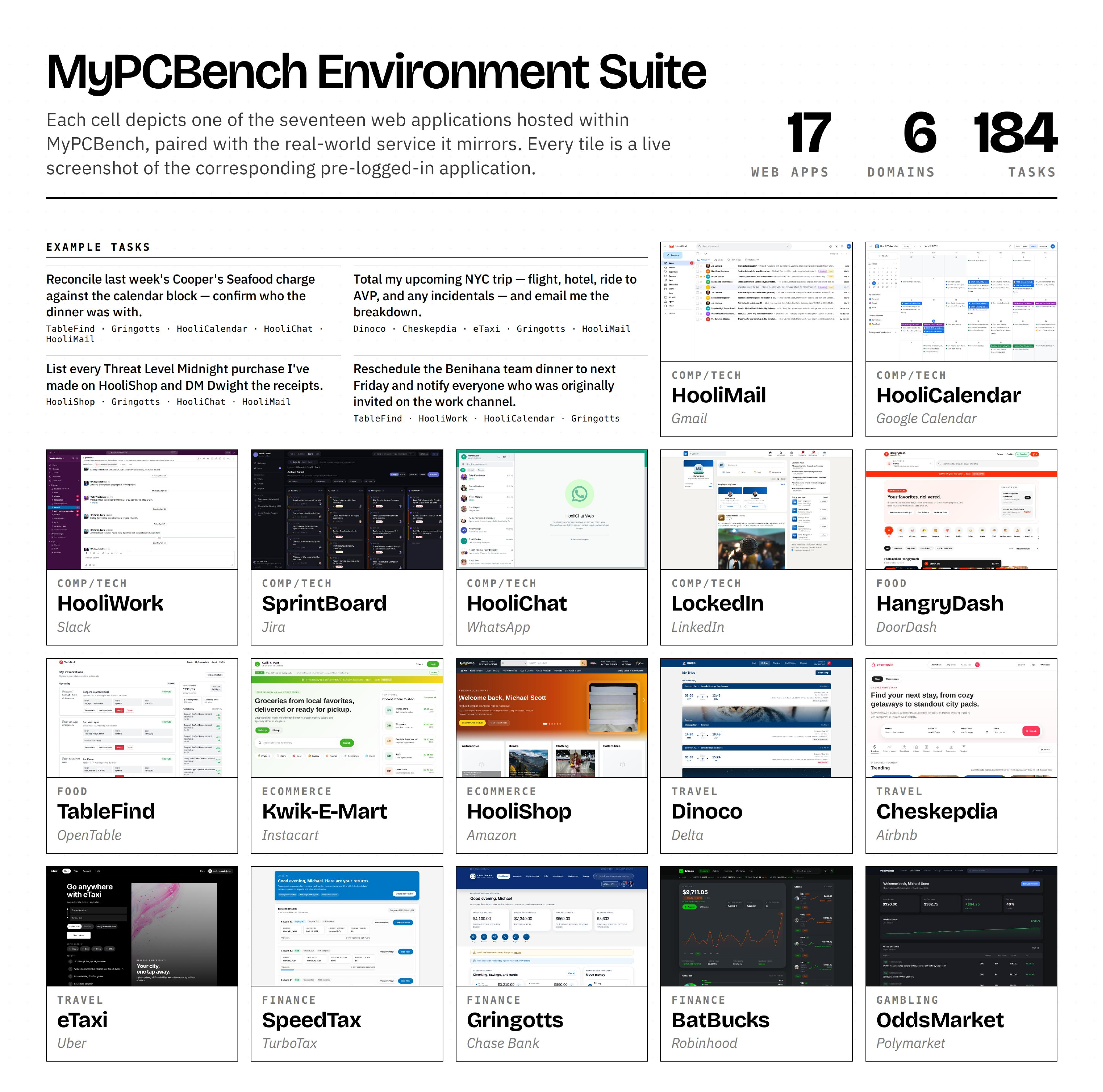}
  \caption{\textbf{\bench{} environment suite.} The \numapps{} logged-in web apps span six SimilarWeb top-level domains (Computers/Tech, Finance, Travel, Food, Ecommerce, Gambling). The four example tasks (top-left) each require threading multiple of these apps against Michael's seeded history.}
  \label{fig:digital-life}
\end{figure}

We release \bench{} as a reproducible, open-source Linux desktop through a Docker image that runs a real QEMU/KVM Ubuntu 24.04 VM with GNOME Shell. The VM hosts \numapps{} pre-logged-in websites (each modeled on a real-world analogue), LibreOffice (Writer, Calc, Impress), and a Firefox profile pre-loaded with a realistic browsing history and bookmark set. Two of the web apps, HooliWork (Slack) and HooliChat (WhatsApp), are also exposed as native desktop apps. The home directory is populated with files relating to Michael's personal and work life. Figure~\ref{fig:digital-life} shows screenshots of all applications. \bench{} is built around three properties for evaluating personalized agents.

\textbf{(1) Cross-app consistency.} Any trip, dinner, or client deal leaves correlated records in every application that would plausibly record it. Michael's Philadelphia trip generates a Cheskepdia (Airbnb) booking, two Gringotts (Chase) charges, a HooliCalendar (Google Calendar) block, two Dinoco (Delta) boarding passes, browsing history for ``Radisson Blu Warwick'', three Travel-folder emails, and HooliChat (WhatsApp) messages referencing the trip. The data seeding pipeline writes those records together so they line up at boot, and runtime cross-app effects keep them in sync. For example, a HangryDash (DoorDash) order posts a charge to Gringotts and drops a confirmation in HooliMail (Gmail).

\textbf{(2) Persona coherence.} The user is a specific person, not a generic account. A real user's friends, co-workers, routines, and preferences are entangled in their data across apps and different data, and our environment preserves that entanglement. Because the persona is \persona{}, we were able to use coding agents to draw on \textit{The Office} canon to populate the environment with large scale coherent, realistic data.

\textbf{(3) Real-world fidelity.} Each web application is a local clone with the security and reproducibility constraints of a fixed VM, but its UI, navigation, and supported flows match the real-world analogue.

\subsection{Environment Creation and Infrastructure}
\label{sec:apps}

\paragraph{Synthetic website generation.} We built \numapps{} clones of real consumer products with Claude Code~\citep{anthropic2026ClaudeCode}, each a full Next.js build rather than a static mock, following prior work on coding-agent web cloning and scaled synthesis of browser environments~\citep{zhou2026wainf,murty2025nnetnav}. The clones implement real workflows. Gringotts supports transfers, bill pay, Zelle, and statement downloads. Dinoco Airlines generates boarding passes with QR codes. eTaxi routes between seeded locations with OSRM, and TableFind exposes a pre-computed reservation inventory of 4{,}128 slots across 31 dates with hold-and-release semantics. Across the canonical \persona{} seed the \numapps{} apps expose 226 distinct database tables and roughly 42{,}000 rows of user-facing state, with the headline record types summarized in Figure~\ref{fig:hero} (transactions, emails, events, messages, orders, browsing history). Per-app catalogs are sized to be browse-realistic, with full counts in Appendix~\ref{app:apps} (Table~\ref{tab:apps}). Product images are real photos (Wikimedia/Wikipedia) rather than synthetic SVGs or non-realistic images, with image--title alignment of every image checked by a vision--language model.\looseness=-1

\paragraph{Persona generation.} The persona is specified as a JSON document covering identity, financial profile, social network, travel history, work context, routines, preferences, and recent and upcoming life events. A deterministic Python pipeline populates every part of the desktop from this spec. It writes SQLite databases for the \numapps{} web apps with cross-consistent references, a Firefox profile with bookmarks/history/cookies/form-fields, and a filesystem of meeting notes, expense reports, trip itineraries, boarding-pass PDFs, and resume drafts.\looseness=-1
\paragraph{Infrastructure.} The default resource budget for a single virtual machine is 4 vCPUs and 8~GB of RAM. Boot-to-ready takes about 90~seconds, and a base snapshot is captured after first boot and used to reset between tasks, avoiding state leakage. The publicly fetchable artifacts are a QEMU wrapper image and a standalone qcow2 disk on the model hub, so evaluators can either run the full guest inside Docker or boot the qcow2 directly under QEMU with no Docker runtime. Each image build runs the full generation pipeline end-to-end, seeding every app database, the Firefox profile, and the user filesystem from the persona spec, before the boot snapshot is taken, so the entire environment is reproducible from the spec alone. Adding personas or websites uses the same template (Appendix~\ref{app:datagen}).

\section{Tasks and Evaluation Setup}
\label{sec:tasks}

\subsection{Task Suite}
\label{sec:cats}

\begin{table}[t]
\centering
\caption{The six behavioral task types in \bench{}, with counts and a representative instruction for each. Full definitions in Appendix~\ref{app:task-types}.}
\label{tab:task-types}
\renewcommand{\arraystretch}{1.04}
\setlength{\tabcolsep}{4pt}
{\footnotesize
\begin{tabular}{@{}p{0.18\textwidth} p{0.1\textwidth} p{0.68\textwidth}@{}}
\toprule
\textbf{Type} & \textbf{\# Tasks (\%)} & \textbf{Representative instruction} \\
\midrule
Bounded action & 64 \mbox{(35\%)} & \textit{Zelle Pam a hundred bucks. She covered me last weekend. Check HooliChat first to make sure Pam Beesly is on my contacts, then put a memo on the transfer.} \\
Multi-step orchestration & 48 \mbox{(26\%)} & \textit{The Threat Level Midnight Fan Club has been dormant. Peek at the group chat, scroll my LockedIn contacts for Dunder Mifflin folks to recruit, draft them an invitation email, and book a watch party on my calendar for next month.} \\
Cross-source reconciliation & 25 \mbox{(14\%)} & \textit{I've got the Jamaica trip AND the Barbados trip booked about four weeks apart. Given my credit-card balance, can I actually afford both, or am I about to max out?} \\
Aggregation \& reporting & 23 \mbox{(12\%)} & \textit{How much am I sending via Zelle each month, and who's getting the money? Check the most recent two complete calendar months and rank the recipients in a LibreOffice Calc spreadsheet.} \\
Personal lookup & 13 \mbox{(7\%)} & \textit{What's my current FlyMiles loyalty tier on Dinoco Airlines, and how many miles do I have in the bank?} \\
Pattern inference & 11 \mbox{(6\%)} & \textit{What do I usually tip on food delivery, in dollars and as a percent? I want to set a smart default so I'm not thinking about it every order.} \\
\bottomrule
\end{tabular}
}
\end{table}

\bench{} includes \numtasks{} tasks, each one inspired by a real use case or request from the OpenClaw community. The authors manually sifted through 2{,}749 anonymized and paraphrased use-cases from the OpenClaw Discord. We dropped requests that (i) were near-duplicates of an already-kept request, (ii) were infeasible inside any deterministic VM (e.g.\ ``call my mom''), or (iii) required an app outside the \numapps{} we host. The remaining requests were rewritten so the named entities (people, restaurants, dates, accounts) match Michael Scott's seeded data. A coding agent generated a per-task rubric in the Odysseys~\citep{jang2026odysseysbenchmarkingwebagents} format. Both the rewrite and the rubric were then audited by the authors (\S\ref{sec:qa}). The final task set is stored as JSON, with each task carrying both its natural-language instruction and its rubric.\looseness=-1

\paragraph{Quality assurance.}\label{sec:qa}
Because coding agents generate the initial task drafts and the application clones, we manually verify both. Each task was reviewed by at least two authors through a custom web interface (Appendix~\ref{app:viewer}, Figure~\ref{fig:reviewer-ui}). Reviewers ran each task end-to-end on the live VM and confirmed that (a)~every named entity exists in the seeded environment, (b)~the expected answer is obtainable from the environment alone, (c)~each rubric criterion is individually checkable from a step-level screenshot, and (d)~the task is not a near-duplicate of another in the suite. All \numtasks{} tasks survived this author QA round.\looseness=-1

\paragraph{Domain coverage.}
We map each application to a top-level SimilarWeb\footnote{\url{https://www.similarweb.com/category/}} category by inspecting its real-world analogue, mirroring the categorization scheme of Odysseys~\citep{jang2026odysseysbenchmarkingwebagents}. The \numapps{} apps span six top-level categories (Computers/Tech, Finance, Travel \& Tourism, Food \& Drink, Ecommerce, Gambling) and fourteen subcategories. The per-app mapping is in Appendix~\ref{app:apps} (Table~\ref{tab:apps}).

\paragraph{Apps per task.}
Tasks span from one to nineteen co-touched applications, 68\% are multi-application, and 40\% span at least two SimilarWeb top-level categories. The multi-application regime is what tests personalization, since the agent has to reconcile data across the persona's environment rather than drive a single tool in isolation. Figure~\ref{fig:task-distributions} (Appendix~\ref{app:task-distributions}) gives the apps-per-task distribution, the per-domain task coverage, and the behavioral task-type split.

\paragraph{Task types.} Each task is also assigned a behavioral \emph{type} that captures what the agent must \emph{do} with the persona's data, independent of which apps are involved or which SimilarWeb domain they fall under (Table~\ref{tab:task-types}). We arrived at the type taxonomy by reading every task instruction and clustering by the primary capability under test.\looseness=-1

\subsection{Agent Harness}
\label{sec:runner}

The harness lets us point standard CUA agents at the \bench{} environment with as little adaptation as possible. We model \bench{} as a partially observable Markov decision process~\citep{kaelbling1998pomdp}. At each step the agent receives an observation (a screenshot of the guest desktop) and emits an action, which the harness exchanges with the guest through the OSWorld-compatible~\citep{xie2024osworld} HTTP Control API on port 5000 (\texttt{GET /screenshot} returns the PNG observation and \texttt{POST /execute} runs the action). VNC/noVNC is exposed only for human observation. Our harness is a thin extension of the OSWorld runner. It boots the \bench{} Docker image, restores a fresh QEMU snapshot before every task so each run begins from an identical desktop state, and drives the standard agent loop until the agent emits \texttt{DONE} or \texttt{FAIL} or exhausts the step budget.\looseness=-1

\paragraph{Observation space.} At each step the agent receives a 1280$\times$800 screenshot of the full Linux desktop, augmented with the action history. Following the OSWorld agent convention, the context keeps the full textual action history but only the 20 most recent screenshots (older ones are replaced by a text placeholder). The Claude agent resizes each screenshot to 1280$\times$720 by configuration. The Qwen agent passes them through the Qwen-VL token budget, and the OpenAI agent passes them as captured.\looseness=-1

\paragraph{Action space.} The action space is the unmodified OSWorld \texttt{pyautogui} surface (click, type, key, scroll, drag, wait, screenshot, done, fail). Each provider's computer-use API defines its own action vocabulary, which we map onto this surface through a translation layer (Claude's \texttt{computer.click} becomes \texttt{click}, CUA's drag path becomes \texttt{drag}). Anthropic's Computer Use bundle~\citep{anthropic2024computeruse} ships a native \texttt{bash} tool and a \texttt{str\_replace\_based\_edit\_tool}. For tool-surface parity we add the vendor-documented shell tool (the OpenAI Responses-API \texttt{shell}, a Qwen function-call \texttt{bash}) to the other agents, so every model runs with the same \texttt{computer}+\texttt{bash} affordance. The OpenAI and Qwen agents also receive one short generic dual-tool hint. Claude receives none, as its native agent already balances the two tools. The \texttt{str\_replace\_based\_edit\_tool} stays Claude-only, as no other provider documents an equivalent. Appendix~\ref{app:cua-bash} gives the controlled cua-only vs.\ cua+bash comparison and Appendix~\ref{app:harness} the full action table, per-provider mapping, system prompts, and per-model action distribution, including when \texttt{bash} helps versus when it triggers the UI-shortcut failure mode of \S\ref{sec:failures}.\looseness=-1

\subsection{Grading}
\label{sec:grading}

We grade each trajectory against its rubric with the full-trajectory-per-rubric LLM-as-a-judge~\citep{zheng2023judgingllmasajudgemtbenchchatbot} of Odysseys~\citep{jang2026odysseysbenchmarkingwebagents}, whose reliability against human judgments was audited and verified. Every task ships a list of natural-language criteria $\{r_1,\dots,r_N\}$ authored alongside it and audited during the QA pass. Rubrics range from 3 to 13 items (mean 6.5 per task), with 1{,}191 in total. The judge runs once per rubric item over the \emph{full} trajectory, receiving the task instruction for context, the single rubric item, the agent's complete action history, and every screenshot from the trajectory in chronological order. Our budget is denominated in \emph{turns}, one LLM call per turn, hard-capped at 100 turns per run. A \emph{step} is one executed action. Because a single turn can emit several \texttt{pyautogui} actions plus a shell call, per-task step counts (and hence the Avg steps in Table~\ref{tab:main-results}) can exceed the turn cap on shell-interleaved cua+bash runs, while Claude emits one action per turn and tops out at 100. We capture one screenshot per turn, so a trajectory never exceeds 100 screenshots. The judge returns ``success'' or ``failure'' per item, $s_{i,r}\in\{0,1\}$, and we let $s_i=\sum_r w_{i,r}\,s_{i,r}$ denote the per-task score, where the authored rubric weights $w_{i,r}$ are normalized to sum to one within each task (most tasks weight criteria unequally to reflect their importance). The judge model is \texttt{gemini-3.1-flash-lite-preview}~\citep{google2026gemini} throughout. 

We report three metrics per model. The \emph{rubric score} $\overline{s}=\frac{1}{T}\sum_i s_i$ is the per-task weighted sum of rubric pass rates (authored weights, normalized per task), then averaged across tasks. It credits partial completion. The stricter \emph{perfect rate} $\frac{1}{T}\sum_i \mathbb{1}[s_i = 1]$ requires every rubric in a task to pass. \emph{Trajectory Efficiency}, also from Odysseys, measures how much rubric score the agent extracts per step,
\[
  \text{Traj.\ Eff.} \;=\; \frac{1}{T}\sum_{i=1}^{T} \frac{s_i}{n_i},
\]
where $n_i$ is the number of agent steps on task $i$. We report it scaled by $100$ (percent of the rubric satisfied per agent step) for readability. The full judge prompt is in Appendix~\ref{app:prompts}.

\section{Experiments and Analysis}
\label{sec:exp}

\subsection{Main Results}
\label{sec:setup}

We evaluate six models on the full \numtasks{}-task suite, each driven by its provider's computer-use (CUA) agent with the shared \texttt{computer}+\texttt{bash} surface and per-provider conventions of \S\ref{sec:runner}. The four closed-weight models are Claude Opus 4.6 and Claude Sonnet 4.6~\citep{anthropic2026opus46,anthropic2026sonnet46}, GPT-5.5~\citep{openai2026gpt55}, and GPT-5.4 mini~\citep{openai2026gpt54mini}. The two open-weight models are Qwen 3.5~\citep{qwen2026qwen35} 35B-A3B and 9B, chosen for contrasting scale within one family. Every run uses the 100-turn budget and shared persona context of \S\ref{sec:grading} and is graded by the same judge. Table~\ref{tab:main-results} reports the three metrics of \S\ref{sec:grading} alongside the average steps each agent consumed.

\begin{table}[t]
\centering
\caption{Main results on the \numtasks{}-task suite under each provider's CUA agent with \texttt{computer}+\texttt{bash} enabled (\texttt{gemini-3.1-flash-lite-preview} judge, 100-turn (LLM-call) budget, shared persona context). \emph{Perfect} is the fraction of tasks for which every rubric in the task passed and is our headline metric. \emph{Rubric score} gives partial credit, and \emph{Traj.\ Eff.} is rubric score per agent step, in percent (all three metrics defined in \S\ref{sec:grading}). cua-only vs.\ cua+bash deltas are in Appendix~\ref{app:cua-bash}.}
\label{tab:main-results}
\renewcommand{\arraystretch}{1.10}
\setlength{\tabcolsep}{6pt}
{\small
\begin{tabular}{@{}lrrrr@{}}
\toprule
\textbf{Model} & \textbf{Perfect $\uparrow$} & \textbf{Rubric score $\uparrow$} & \textbf{Avg steps} & \textbf{Traj.\ Eff.\ $\uparrow$} \\
\midrule
\multicolumn{5}{@{}l}{\textit{API (closed-weight)}} \\
\quad Claude Opus 4.6 & \textbf{55.4} & \textbf{81.8} & 46.5 & \textbf{3.61} \\
\quad Claude Sonnet 4.6 & 39.1 & 65.4 & 45.8 & 3.03 \\
\quad GPT-5.5           & 29.3 & 54.1 & 45.8 & 1.45 \\
\quad GPT-5.4 mini      & 19.0 & 48.8 & 43.7 & 1.65 \\
\midrule
\multicolumn{5}{@{}l}{\textit{Open weights}} \\
\quad Qwen 3.5 35B-A3B  & \phantom{0}7.6 & 42.5 & 66.0 & 1.41 \\
\quad Qwen 3.5 9B       & \phantom{0}2.7 & \phantom{0}7.0 & 69.2 & 0.65 \\
\bottomrule
\end{tabular}
}
\end{table}

Closed-weight frontier agents lead by a wide margin. Claude Opus 4.6 reaches \textbf{55.4\%} perfect at 81.8\% rubric score, the only model above 50\%, 1.4$\times$ the next-best (Claude Sonnet 4.6, 39.1\%) and nearly twice the best non-Claude model (GPT-5.5, 29.3\%). Within the open-weight tier, Qwen 3.5 35B-A3B nearly triples 9B on perfect rate (7.6 vs.\ 2.7), and the 9B collapses under the dual-tool surface (20.2$\to$7.0 rubric against its cua-only baseline, with the breakdown in Appendix~\ref{app:cua-bash}). Each cell is a single canonical run per task, so we read the large cross-model gaps rather than small per-cell differences.

Trajectory Efficiency adds a step-budget view. Opus extracts \textbf{3.61} rubric points per step, over 5$\times$ Qwen 3.5 9B (0.65). With \texttt{bash} enabled the OpenAI and Qwen agents interleave shell calls with GUI actions and spend more steps per task than their cua-only baselines, leaving GPT-5.5 and GPT-5.4 mini at 1.45 and 1.65. Step count alone does not predict efficiency. A similar count can reflect tight execution (Sonnet, 45.8 steps, Eff.\ 3.03) or unproductive looping (Qwen 9B, 69.2 steps, Eff.\ 0.65). The failure-mode breakdown in \S\ref{sec:failures} separates the two.\looseness=-1

\subsection{Performance by Task Type}
\label{sec:tasktype}

\begin{figure}[t]
\centering
\begin{minipage}[t]{0.575\textwidth}
\centering
\includegraphics[width=\linewidth]{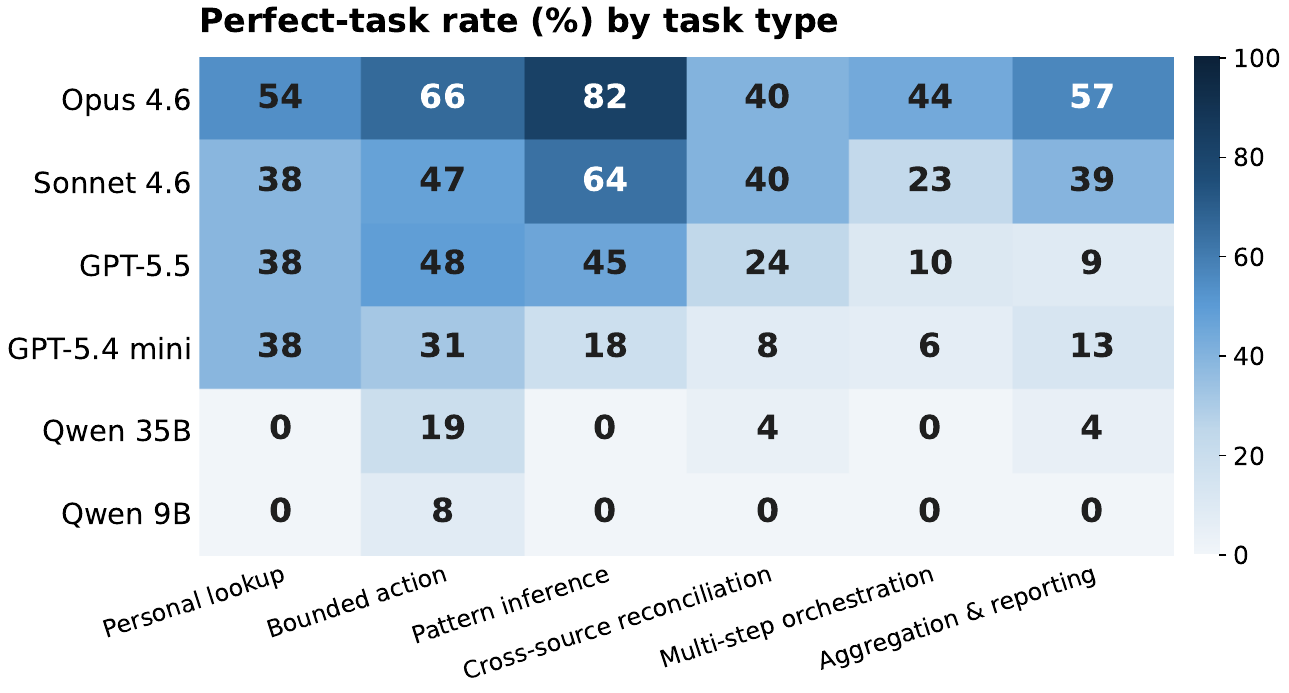}
\end{minipage}\hfill
\begin{minipage}[t]{0.405\textwidth}
\centering
\includegraphics[width=\linewidth]{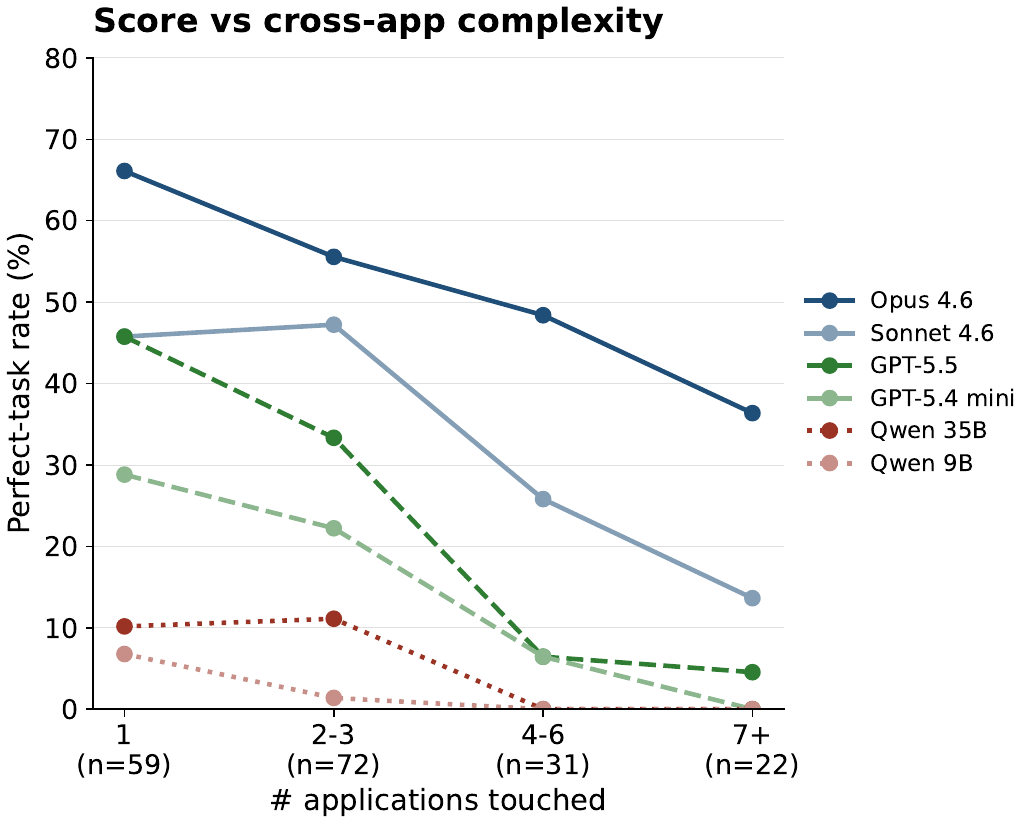}
\end{minipage}
\caption{\textbf{Left:} per-task-type perfect rate (\%), models $\times$ task types (all models with \texttt{computer}+\texttt{bash}). \textbf{Right:} perfect rate versus the number of distinct apps a task touches. At 7+ apps, only the Claude tier and GPT-5.5 (4.5\%) perfect any task.}
\label{fig:results-breakdown}
\end{figure}

Figure~\ref{fig:results-breakdown}-left shows the per-type perfect rate for every model.

Two categories localize the gap. On \emph{personal lookup} every model in the API tier clears 38\% perfect (Opus leads at 54\%). On \emph{bounded action} only Opus, Sonnet, and GPT-5.5 stay above 46\%. The remaining four categories all require reasoning over persona history or coordinating writes across multiple apps, and the gap to Opus widens accordingly. The gap stays wide on \emph{pattern inference} even with \texttt{bash} enabled. Opus reaches 82\% perfect and GPT-5.5 45\% on the same 11 tasks. These tasks ask the agent to infer an unstated rule from many records (``what do I usually tip?''), and the rubric only credits answers that match the rule the seeded history supports. On \emph{aggregation} and \emph{multi-step orchestration} the OpenAI CUA family and both Qwen models stay below 16\% perfect (GPT-5.5 recovers \emph{cross-source reconciliation} to 24\%), and Qwen 9B perfects zero tasks across all four analysis categories (personal lookup, aggregation, pattern inference, cross-source reconciliation).

\subsection{Performance Scaling by Steps and Apps}
\label{sec:horizon}

We study how performance scales along two axes, the number of distinct applications a task touches and the number of agent steps the trajectory consumes (Figure~\ref{fig:perfect-vs-steps} and Table~\ref{tab:apps-scaling}, Appendix~\ref{app:results-tables}).

\paragraph{Apps and steps both stress horizon.} From single-app to 7+-app bins the perfect rate falls from 66\% to 36\% for Opus and 46\% to 14\% for Sonnet, while GPT-5.4 mini, Qwen 35B, and Qwen 9B all reach 0\% at 7+ apps and GPT-5.5 reaches only 4.5\% (per-bin perfect rates in Figure~\ref{fig:results-breakdown}-right and rubric scores in Table~\ref{tab:apps-scaling}). On the step axis (Figure~\ref{fig:perfect-vs-steps}) Opus is still climbing at the 100-step cap, GPT flattens by step 60, and Qwen saturates by step 25.\looseness=-1

\begin{figure}[!t]
\centering
\includegraphics[width=\textwidth,height=0.45\textheight,keepaspectratio]{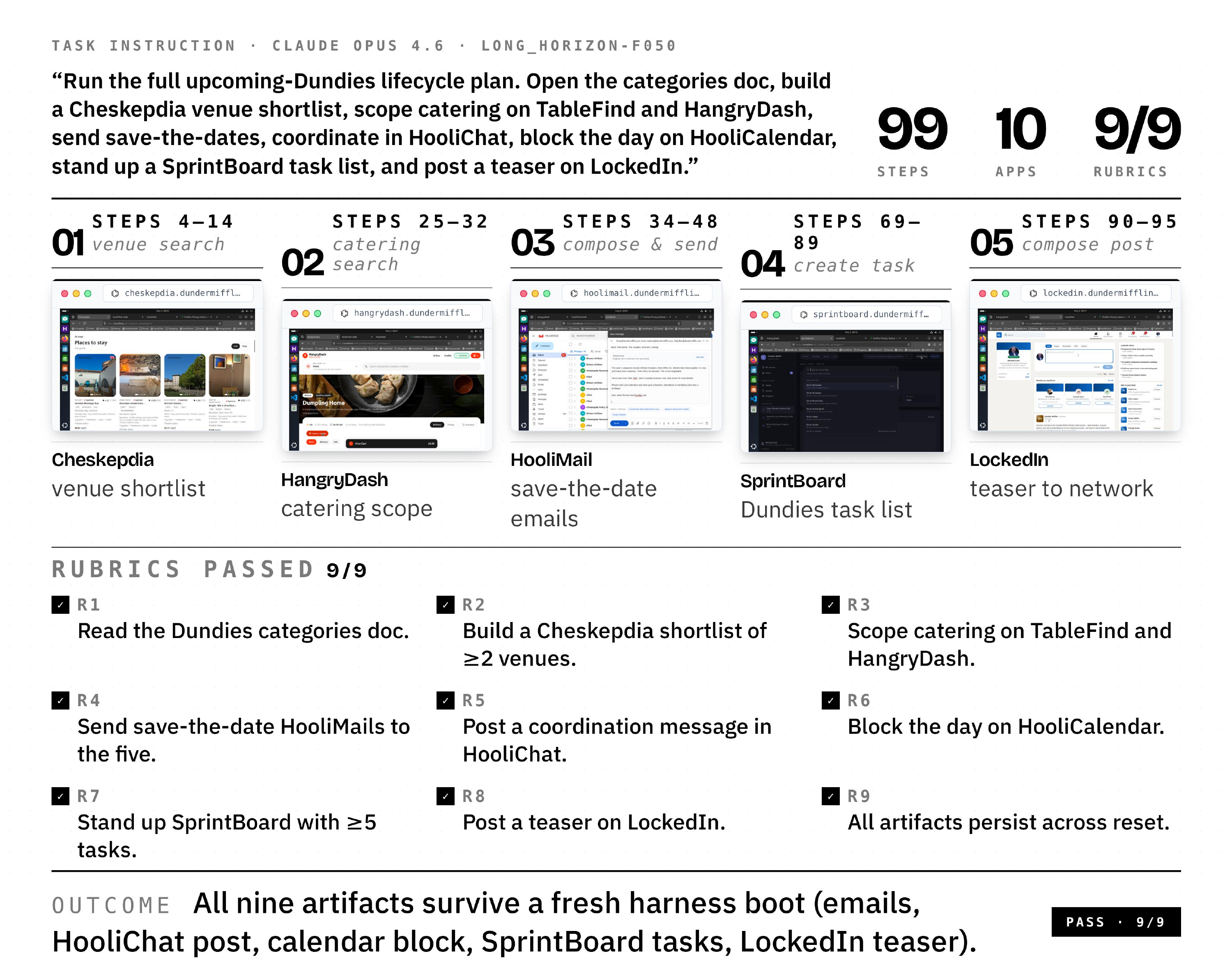}
\caption{\textbf{One full successful Opus trajectory, the Dundies-lifecycle plan on \texttt{long\_horizon-f050}} (99 steps, 10 apps, 9/9 rubrics). Cells are real screenshots from the steps where the agent is actively driving each app. The bottom strip enumerates the nine rubric criteria the judge marked passed.}
\label{fig:walkthrough}
\end{figure}

\subsection{Personalization-Specific Failures}
\label{sec:failures}

We classify every failed-rubric judge explanation into one of five modes (Appendix~\ref{app:failures-detail}). Premature \texttt{DONE} (354 hits) and skipped required app (323) account for most of the loss, followed by surface-error abandonment (129), partial artifact (47), and hallucinated persona data (31). The three families concentrate in different modes. GPT dominates premature \texttt{DONE} (235 of 354 hits), Qwen drives persona-data hallucination (13 of 31), and Claude takes console-script shortcuts, driving via \texttt{bash} rather than the UI even though all models now have that tool. The action distributions in Appendix~\ref{app:harness} confirm the split. Zero-score trajectories split into two regimes. The GPT family and Sonnet abandon early (mean 22--31 steps), while Opus and the Qwen models keep working past the point where the rubric is recoverable (mean 52--85 steps). Figure~\ref{fig:walkthrough} traces a clean Opus run for contrast, and Appendix~\ref{app:trajectories} gives per-family pass/fail trajectories visualized.\looseness=-1

\section{Conclusion}
\label{sec:conclusion}

Our benchmark provides improvement points for every model evaluated in the lens of personally intelligent computer-use agents. The gaps on \bench{} resolve into three family-shaped failure patterns (\S\ref{sec:failures}). Claude shortcuts past the UI through \texttt{bash}, the GPT family premature-\texttt{DONE}s before the rubric-graded side-effect, and within Qwen the 35B hallucinates persona values while the 9B collapses under the dual-tool schema. These are not just aggregate-rate gaps but specific modes that future agent designs can target. The best agent perfects barely more than half the suite, the cross-app and long-horizon slopes are steep for every other model, and the hardest categories (aggregation \& reporting, multi-step orchestration, pattern inference) remain the weakest. Additionally, we find that the ability for models to leverage their coding abilities in personal assistant capabilities while balancing with computer-use actions will be very important going forward, with gaps showing currently. We hope that \bench{} pushes computer-use agents and general personal assistant research towards a personalized lens, where we believe there is currently a large gap between deployment and evaluation. We release the \bench{} environment, tasks with rubrics, harness, and rubric-grading judge (\url{https://mypcbench.com}) as a baseline for work on personal computer-use agents.\looseness=-1

\bibliographystyle{plainnat}
\bibliography{references}

\appendix

\let\OldAppendixSection\section
\renewcommand{\section}[1]{\FloatBarrier\OldAppendixSection{#1}}

\section{Application Details}
\label{app:apps}

Table~\ref{tab:apps} summarizes the \numapps{} web applications hosted within the \bench{} environment image, the real-world service each one mirrors, and the SimilarWeb top-level category and subcategory inherited from that analogue.

\begin{table}[t]
  \centering
  \caption{The \numapps{} web applications hosted within the \bench{} environment image, with the SimilarWeb top-level category and subcategory each one inherits from its real-world analogue.  Screenshots of every app are in Figure~\ref{fig:digital-life}.}
  \label{tab:apps}
  \renewcommand{\arraystretch}{1.08}
  \setlength{\tabcolsep}{3pt}
  {\footnotesize
  \begin{tabular}{@{}p{0.115\textwidth} p{0.10\textwidth} p{0.18\textwidth} p{0.16\textwidth} p{0.34\textwidth}@{}}
  \toprule
  \textbf{App} & \textbf{Analogue} & \textbf{SimilarWeb category} & \textbf{Subcategory} & \textbf{Description} \\
  \midrule
  HooliMail & Gmail & Computers, Electronics \& Tech & Email & Gmail-style web client over a local Maildir, with 2{,}398 seeded messages across Inbox, Sent, and four labeled folders. \\
  HooliCalendar & Google Calendar & Computers, Electronics \& Tech & Productivity & Google-Calendar-style scheduling app exposing 679 personal and work events with recurrence and attendee lists. \\
  HooliWork & Slack & Computers, Electronics \& Tech & Programming \& Developer Software & Slack-style team messenger with the persona's branch channels, DMs, and read state. \\
  SprintBoard & Jira & Computers, Electronics \& Tech & Programming \& Developer Software & Jira-style issue tracker holding the persona's running sprints, tickets, and assignees. \\
  HooliChat & WhatsApp & Computers, Electronics \& Tech & Telecommunications & WhatsApp-style messenger with one-to-one and group threads spanning friends, family, and co-workers. \\
  LockedIn & LinkedIn & Computers, Electronics \& Tech & Social Media Networks & LinkedIn-style professional network exposing the persona's profile, feed, and connections, with 18 jobs across 9 companies and 13 fully populated profiles, plus connection- and company-authored feed posts on top of the user's own. \\
  HangryDash & DoorDash & Food \& Drink & Restaurants \& Delivery & DoorDash-style food delivery surface covering 28 restaurants with 165 menu items, plus order history and active carts. \\
  TableFind & OpenTable & Food \& Drink & Restaurants \& Delivery & OpenTable-style restaurant reservation app over a 4{,}128-slot pre-computed inventory across 31 dates and 19 restaurants, with past bookings and search. \\
  Kwik-E-Mart & Instacart & Ecommerce \& Shopping & Marketplace & Instacart-style grocery delivery spanning 7 Scranton-area stores (Wegmans, Price Chopper, ALDI, Gerrity's, Weis, Trader Joe's, Giant) with 109 curated SKUs total, multi-store carts, and an order log. \\
  HooliShop & Amazon & Ecommerce \& Shopping & Marketplace & Amazon-style retail front with a 90-SKU catalog of real-photo product imagery, personalized recommendations, orders, and a cart. \\
  Dinoco & Delta & Travel \& Tourism & Air Travel & Delta-style airline app exposing flights, loyalty, baggage, and check-in, with 14 booked flights across 5 AVP hubs and full real-Delta seat parity, plus a dynamic alternative-flight search pool. \\
  Cheskepdia & Airbnb & Travel \& Tourism & Accommodation \& Hotels & Airbnb-style stays / experiences booking surface with trips, wishlists, and search. \\
  eTaxi & Uber & Travel \& Tourism & Ground Transportation & Uber-style ride-hail app with ride history, saved places, and active requests. \\
  SpeedTax & TurboTax & Finance & Accounting \& Auditing & TurboTax-style tax preparation surface with prior-year returns, W-2s, 1099s, and a current-year draft. \\
  Gringotts & Chase Bank & Finance & Banking, Credit \& Lending & Chase-style bank dashboard covering checking, savings, credit card, and a 1{,}812-row transaction log. \\
  BatBucks & Robinhood & Finance & Investing & Robinhood-style brokerage with a holdings view, watchlist, and position-level history. \\
  OddsMarket & Polymarket & Gambling & Other & Polymarket-style prediction-market exchange with the persona's open positions and watchlist. \\
  \bottomrule
  \end{tabular}
  }
\end{table}

\section{Task-Distribution Plots}
\label{app:task-distributions}

\begin{figure}[h]
\centering
\includegraphics[width=\textwidth]{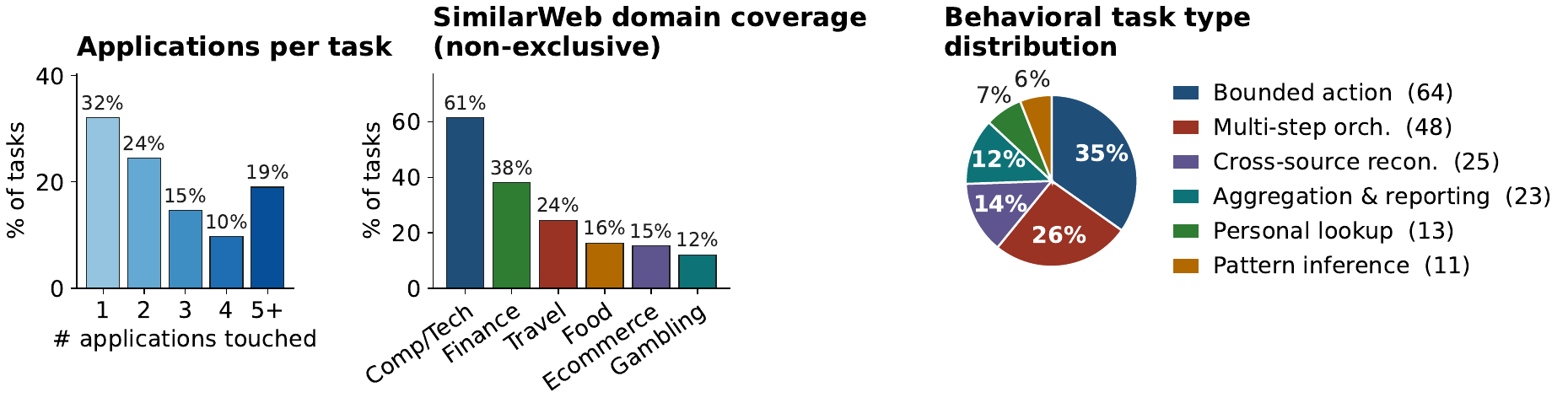}
\caption{\textbf{Left:} distribution of tasks by the number of distinct applications they touch. \textbf{Middle:} fraction of tasks that touch at least one application in each SimilarWeb top-level category (non-exclusive, so a single multi-app task can contribute to several bars). \textbf{Right:} behavioral task-type split (exclusive 1-of-6 categorization per task).}
\label{fig:task-distributions}
\end{figure}

\section{Task-Type Definitions}
\label{app:task-types}

\begin{table}[h]
\centering
\caption{The six behavioral task types in \bench{}, with definition, count, and a representative instruction for each. Counts cover all \numtasks{} tasks, and the per-task type mapping ships with the release as \texttt{tasks/final/task\_types.json}. (Main paper Table~\ref{tab:task-types} reproduces only the counts and example instructions.)}
\label{tab:task-types-app}
\renewcommand{\arraystretch}{1.15}
\setlength{\tabcolsep}{4pt}
{\footnotesize
\begin{tabular}{@{}p{0.135\textwidth} p{0.06\textwidth} p{0.36\textwidth} p{0.36\textwidth}@{}}
\toprule
\textbf{Type} & \textbf{n (\%)} & \textbf{Definition} & \textbf{Representative instruction} \\
\midrule
Bounded action & 64 \mbox{(35\%)} & Execute one concrete write action, or a tightly-scoped sequence within a single application, grounded in the persona's current state. & \textit{Zelle Pam a hundred bucks. She covered me last weekend. Check HooliChat first to make sure Pam Beesly is on my contacts, then put a memo on the transfer.} \\
\midrule
Multi-step orchestration & 48 \mbox{(26\%)} & Chain reads and writes across multiple applications, typically producing artifacts (LibreOffice docs, decks) plus coordinated side-effects (calendar blocks, sent messages, board updates). & \textit{The Threat Level Midnight Fan Club has been dormant. Peek at the group chat, scroll my LockedIn contacts for Dunder Mifflin folks to recruit, draft them an invitation email, and book a watch party on my calendar for next month.} \\
\midrule
Cross-source reconciliation & 25 \mbox{(14\%)} & Reconcile a claim, assumption, or hypothetical against the persona's data by pulling from several apps and quantifying the gap. Covers both contradiction-finding and counterfactual feasibility. & \textit{I've got the Jamaica trip AND the Barbados trip booked about four weeks apart. Given my credit-card balance, can I actually afford both, or am I about to max out?} \\
\midrule
Aggregation \& reporting & 23 \mbox{(12\%)} & Compute a total, distribution, ranking, chart, or rollup from many persona records, typically delivered into a LibreOffice document. & \textit{How much am I sending via Zelle each month, and who's getting the money? Check the most recent two complete calendar months and rank the recipients in a LibreOffice Calc spreadsheet.} \\
\midrule
Personal lookup & 13 \mbox{(7\%)} & Surface a specific named value, file, or record from the persona's environment. Single-fact retrieval, no rollup, no inference. & \textit{What's my current FlyMiles loyalty tier on Dinoco Airlines, and how many miles do I have in the bank?} \\
\midrule
Pattern inference & 11 \mbox{(6\%)} & Infer an unstated habit, preference, or stylistic pattern from historical persona data, never explicitly stored anywhere in the environment. & \textit{What do I usually tip on food delivery, in dollars and as a percent? I want to set a smart default so I'm not thinking about it every order.} \\
\bottomrule
\end{tabular}
}
\end{table}

\section{Tool Surface: cua-only vs.\ cua+bash}
\label{app:cua-bash}

The main results (Table~\ref{tab:main-results}) evaluate every model with both the \texttt{computer} and \texttt{bash} tools. Anthropic's native computer-use agent already exposes \texttt{bash} and a file editor, so the Claude rows are unchanged from a computer-only configuration. For OpenAI and Qwen we add the vendor-documented \texttt{bash}/\texttt{shell} tool to the otherwise computer-only agent.

\paragraph{Dual-tool usage hint.} A computer-only agent given a shell tool with no guidance tends to derive an answer from the shell and stop, without performing the requested workflow in the GUI (the ``premature \texttt{DONE}'' mode of \S\ref{sec:failures}). The OpenAI agents therefore receive a short benchmark-agnostic block appended to the system prompt (\texttt{GUI\_WORKFLOW\_HINT} in \texttt{agents/prompts.py}).

\begin{quote}\small\itshape
When you have both a visual/GUI tool and a shell/terminal tool, treat them as complementary. Use shell for read-only work --- inspecting files, querying local data, parsing or computing. Use the GUI tool to actually perform whatever the user asked you to do in the visible environment. Don't substitute shell exploration for visible action: producing an answer in your text response without performing the requested workflow visibly typically leaves the task incomplete.
\end{quote}

The Qwen cua+bash agent receives the equivalent guidance inside its bash tool description (\texttt{\_BASH\_TOOL\_DESCRIPTION} in \texttt{agents/qwen\_cua.py}), which reads ``Use bash for read-only data work --- file inspection (\texttt{ls}, \texttt{cat}, \texttt{find}), querying local SQLite DBs, parsing or computing over text. Use the GUI tool to actually perform whatever the user asked you to do in the visible environment; don't substitute shell exploration for visible action.''

Claude receives no such hint. Its native agent already balances the two tools (the action distributions in Appendix~\ref{app:harness} show a stable bash/computer mix without prompting). This is the one prompt asymmetry between families. Without the hint, GPT-5.5 with \texttt{bash} \emph{regresses} by roughly 24 rubric points relative to its computer-only baseline (measured on a preliminary full run, not shown in Table~\ref{tab:cua-bash}), so the hint is what makes ``GPT + documented shell'' a fair comparison rather than a sandbagged one.

\begin{table}[h]
\centering
\caption{Effect of adding the documented \texttt{bash}/\texttt{shell} tool to the computer-only (cua-only) OpenAI and Qwen agents, on the full \numtasks{}-task suite (identical persona context, 100-turn (LLM-call) budget, same Gemini judge). cua-only is the computer-only baseline on the same image. Every cell (both modes) is a single canonical run per task with no best-of-N or score-maximizing selection, and cua+bash adds \texttt{bash} plus the dual-tool hint above. $\Delta$ is cua+bash minus cua-only. The Qwen 3.5 9B rubric drop ($-13.2$) is the one large delta. The GPT-5.5, GPT-5.4 mini, and Qwen 35B rubric deltas and \emph{all} perfect-count deltas are small relative to the cross-model gaps in Table~\ref{tab:main-results} and are reported for context.}
\label{tab:cua-bash}
\renewcommand{\arraystretch}{1.18}
\setlength{\tabcolsep}{6pt}
{\small
\begin{tabular}{@{}lrrrrrr@{}}
\toprule
 & \multicolumn{2}{c}{\textbf{Perfect}} & \multicolumn{2}{c}{\textbf{Rubric score}} & \multicolumn{2}{c}{\textbf{$\Delta$}} \\
\cmidrule(lr){2-3}\cmidrule(lr){4-5}\cmidrule(lr){6-7}
\textbf{Model} & \textbf{cua-only} & \textbf{cua+bash} & \textbf{cua-only} & \textbf{cua+bash} & \textbf{Perfect} & \textbf{Rubric} \\
\midrule
GPT-5.5          & 23.9 & 29.3 & 50.5 & 54.1 & $+5.4$ & $+3.6$ \\
GPT-5.4 mini     & 18.5 & 19.0 & 47.5 & 48.8 & $+0.5$ & $+1.3$ \\
Qwen 3.5 35B-A3B & 10.3 & \phantom{0}7.6 & 39.0 & 42.5 & $-2.7$ & $+3.5$ \\
Qwen 3.5 9B      & \phantom{0}4.3 & \phantom{0}2.7 & 20.2 & \phantom{0}7.0 & $-1.6$ & $-13.2$ \\
\bottomrule
\end{tabular}
}
\end{table}

The clearest effect in Table~\ref{tab:cua-bash} is that \textbf{Qwen 3.5 9B regresses sharply} when given \texttt{bash} ($-13.2$ rubric). Inspection of its trajectories shows it routinely emits malformed tool calls that splice the \texttt{bash} and \texttt{computer} schemas together, so a documented tool is not a free affordance and below a capability threshold is actively harmful. The GPT-5.5, GPT-5.4 mini, and Qwen 35B differences are small relative to the cross-family gaps in Table~\ref{tab:main-results}. We therefore make no strong claim that adding \texttt{bash} helps or hurts those three models, only that equalizing the tool surface does not overturn the model ordering.

\section{Per-Task-Type, Cross-App, and Step-Budget Scaling}
\label{app:results-tables}

\begin{figure}[h]
\centering
\includegraphics[width=\textwidth]{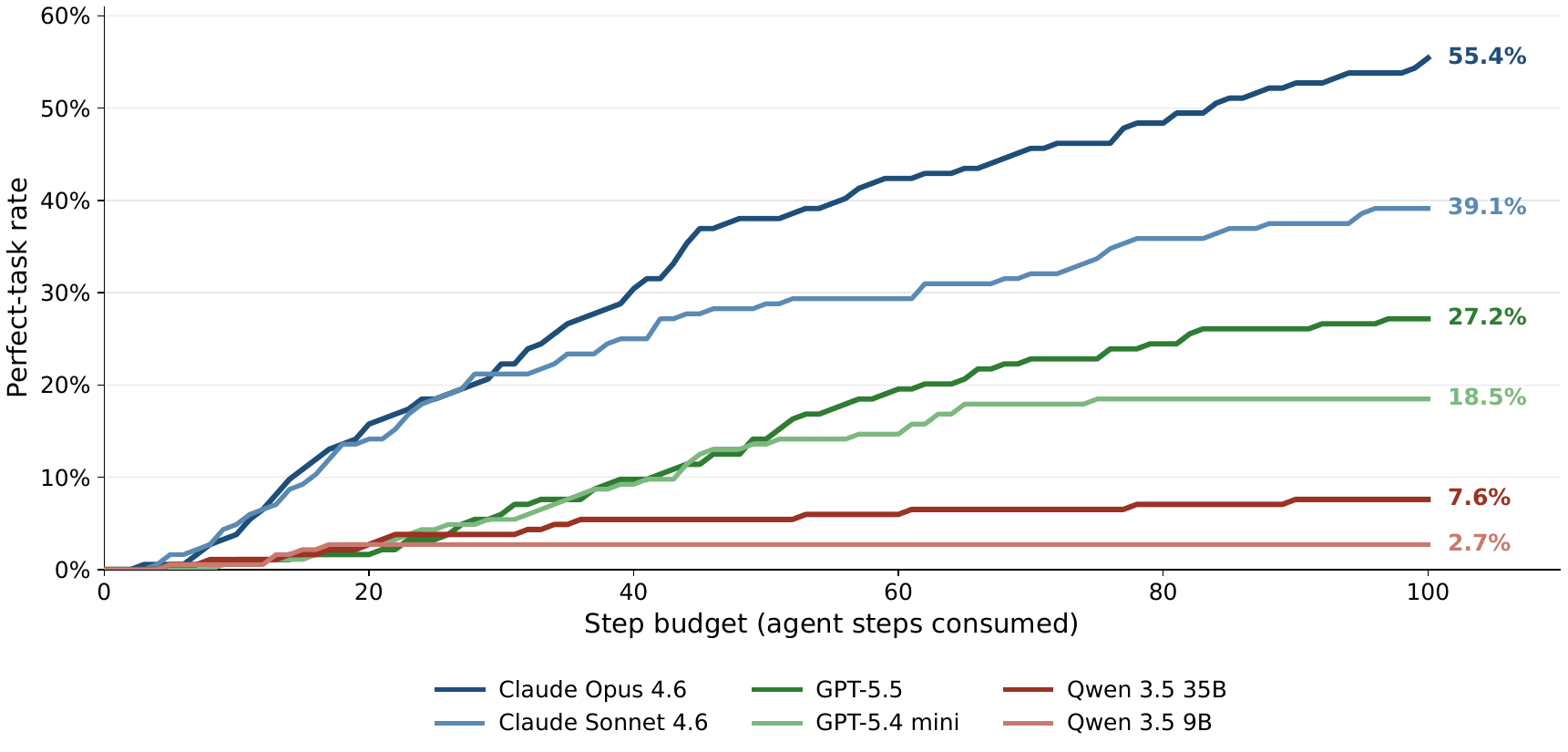}
\caption{\textbf{Step-budget scaling law.} For each model, the curve at step budget $X$ is the fraction of the \numtasks{} tasks the model graded perfect with $\le\!X$ agent steps consumed. Because cua+bash step counts can exceed the 100-turn budget (\S\ref{sec:grading}), perfect tasks completed in more than 100 steps fall outside the plotted range, so a curve can terminate below the model's Table~\ref{tab:main-results} perfect rate (GPT-5.5 ends at 27.2 versus 29.3 overall for this reason). Curve shape, not just height, separates the families.}
\label{fig:perfect-vs-steps}
\end{figure}

Tables~\ref{tab:by-type} and~\ref{tab:apps-scaling} are the raw numbers behind Figure~\ref{fig:results-breakdown}. Figure~\ref{fig:perfect-vs-steps} reads the step axis as a scaling law.

\begin{table}[h]
\centering
\caption{Rubric score (\%) by task type. Rows ordered by descending cross-model average. \emph{Mean} is the unweighted average across the six models. Best per row in \textbf{bold}.}
\label{tab:by-type}
\renewcommand{\arraystretch}{1.10}
\setlength{\tabcolsep}{4pt}
{\footnotesize
\begin{tabular}{@{}lrrrrrrrr@{}}
\toprule
\textbf{Task type} & \textbf{n} & \textbf{Opus} & \textbf{Sonnet} & \textbf{GPT-5.5} & \textbf{GPT-5.4 mini} & \textbf{Qwen 35B} & \textbf{Qwen 9B} & \textbf{Mean} \\
\midrule
Bounded action               & 64 & \textbf{85.3} & 70.4 & 73.6 & 63.6 & 58.9 & 14.4 & 61.0 \\
Pattern inference            & 11 & \textbf{94.7} & 77.3 & 59.1 & 53.2 & 35.9 & \phantom{0}6.1 & 54.4 \\
Personal lookup              & 13 & \textbf{83.8} & 66.0 & 61.2 & 58.5 & 33.2 & 19.2 & 53.7 \\
Cross-source reconciliation  & 25 & \textbf{76.1} & 61.8 & 57.7 & 52.4 & 37.6 & \phantom{0}1.3 & 47.8 \\
Aggregation \& reporting     & 23 & \textbf{81.9} & 61.9 & 34.5 & 47.6 & 24.7 & \phantom{0}0.6 & 41.9 \\
Multi-step orchestration     & 48 & \textbf{76.6} & 59.5 & 32.5 & 24.3 & 35.6 & \phantom{0}0.0 & 38.1 \\
\bottomrule
\end{tabular}
}
\end{table}

\begin{table}[h]
\centering
\caption{Rubric score (\%) versus the number of distinct applications a task touches. 68\% of \bench{} tasks are multi-app.}
\label{tab:apps-scaling}
\renewcommand{\arraystretch}{1.10}
\setlength{\tabcolsep}{6pt}
{\footnotesize
\begin{tabular}{@{}lrrrrrrr@{}}
\toprule
\textbf{Apps touched} & \textbf{n} & \textbf{Opus} & \textbf{Sonnet} & \textbf{GPT-5.5} & \textbf{GPT-5.4 mini} & \textbf{Qwen 35B} & \textbf{Qwen 9B} \\
\midrule
1   & 59 & 87.4 & 69.9 & 67.3 & 58.2 & 44.8 & 14.5 \\
2--3 & 72 & 82.4 & 66.8 & 63.1 & 56.5 & 49.0 & \phantom{0}5.7 \\
4--6 & 31 & 79.8 & 61.8 & 32.6 & 36.2 & 33.0 & \phantom{0}0.5 \\
7+   & 22 & 67.9 & 54.1 & 19.5 & 16.6 & 28.3 & \phantom{0}0.0 \\
\midrule
$\Delta$ (1 $\rightarrow$ 7+) & & $-19.5$ & $-15.8$ & $-47.8$ & $-41.6$ & $-16.5$ & $-14.5$ \\
\bottomrule
\end{tabular}
}
\end{table}

\section{Family-Signature Plots}
\label{app:family-detail}

\begin{figure}[h]
\centering
\includegraphics[width=\textwidth]{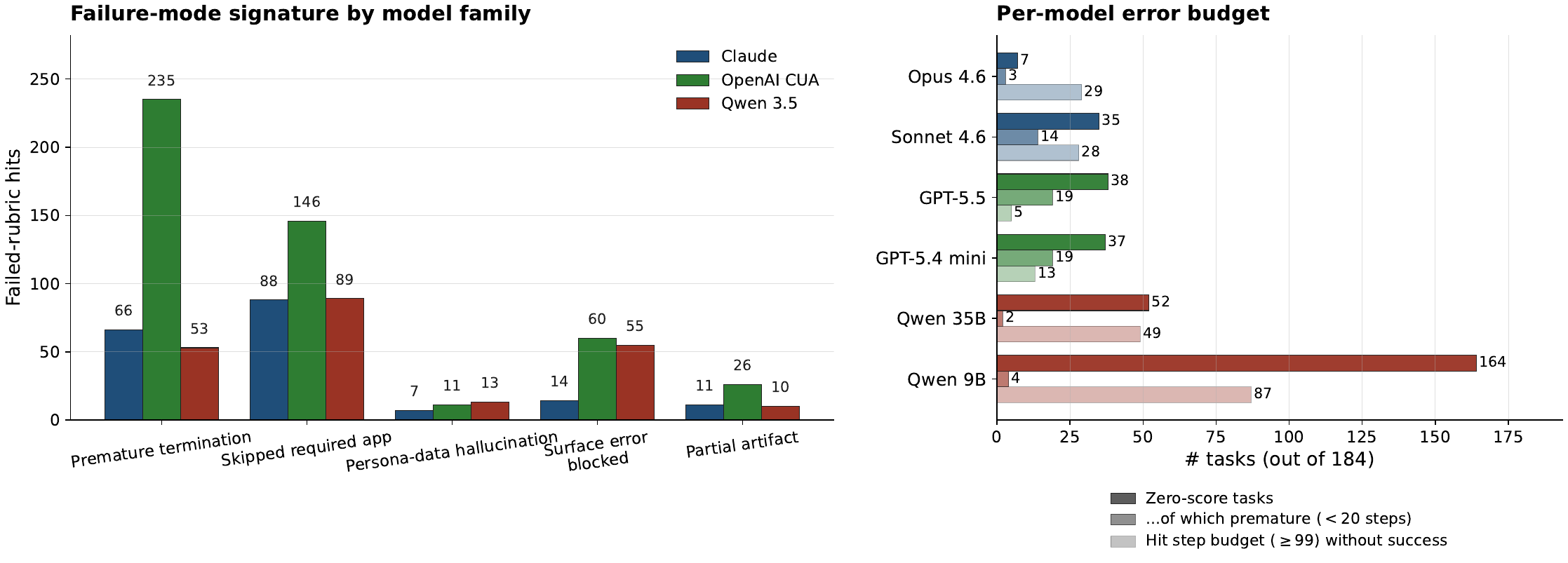}
\caption{\textbf{Left:} failed-rubric hits by failure mode, grouped by family (Claude = Opus 4.6 + Sonnet 4.6, OpenAI CUA = GPT-5.5 + GPT-5.4 mini, Qwen = 3.5 35B-A3B + 9B). \textbf{Right:} per-model error budget. Top dark bar: zero-score tasks. Middle: subset that terminated under 20 steps. Lightest: trajectories that hit the step budget ($\geq$99 steps) without success.}
\label{fig:family-signature}
\end{figure}

Figure~\ref{fig:family-signature} breaks failed-rubric hits out by failure mode and groups them by family (left), and shows a per-model error budget (right).

\section{Detailed Failure Modes}
\label{app:failures-detail}

\begin{table}[h]
\centering
\caption{Failure-mode counts on rubric items the judge marked failed, aggregated across all six models. Per-family breakouts are in Figure~\ref{fig:family-signature}.}
\label{tab:failure-modes}
\renewcommand{\arraystretch}{1.05}
\setlength{\tabcolsep}{6pt}
{\footnotesize
\begin{tabular}{@{}lrp{0.55\textwidth}@{}}
\toprule
\textbf{Failure mode} & \textbf{Hits} & \textbf{Description} \\
\midrule
Premature termination       & 354 & Agent emits \texttt{DONE} before the trajectory satisfies the remaining rubric items. \\
Skipped required app        & 323 & Multi-app task closed after only some apps are visited, so the unvisited app's rubrics fail. \\
Surface error as terminal   & 129 & Agent hits a captcha, console error, slow page, or modal and quits instead of recovering. \\
Partial artifact            &  47 & Artifact started but not saved/exported (e.g., spreadsheet opened but never saved). \\
Hallucinated persona data   &  31 & Agent fabricates a value instead of reading the seeded source. \\
\bottomrule
\end{tabular}
}
\end{table}

\paragraph{Why the perfect rate falls faster than the rubric score.} A failed rubric is rarely an isolated event. Skipped-app failures co-occur with premature-\texttt{DONE} on the same trajectory (the agent quits because it considers itself done after the apps it did open), and surface errors trigger partial-artifact failures (an opened spreadsheet that is never saved). Because perfect rate requires \emph{every} rubric to pass, even one such co-occurring failure zeroes the task.

\paragraph{Per-family breakdown.} Three trends fall out of the per-model counts in Table~\ref{tab:failure-modes} and Figure~\ref{fig:family-signature}.
\begin{itemize}[leftmargin=*,nosep]
  \item \textbf{The GPT family stops too early.} The GPT-family concentration of premature-\texttt{DONE} hits noted in \S\ref{sec:failures} splits as GPT-5.4 mini 130 and GPT-5.5 105, more than 6$\times$ either Claude model on its own (Opus 28, Sonnet 38).
  \item \textbf{The Qwen family shows different errors.} Qwen 35B drives the family's qualitative failures, with 13 of the 31 persona-data-hallucination hits and 55 of the 129 surface-error abandonments (43\%), the most of any single model in both modes (versus 7 hallucination hits in Claude and 11 in GPT). Qwen 9B fails differently --- it cannot maintain the dual \texttt{computer}+\texttt{bash} tool schema and zero-scores 164 of 184 tasks (rubric mean 7.0), collapsing before its trajectories accumulate a classifiable failure explanation at all.
  \item \textbf{The Claude family takes UI shortcuts.} All six models now have the same \texttt{bash} tool (Appendix~\ref{app:cua-bash}), and the GPT family actually uses it \emph{more} in raw volume (52\%/44\% of actions vs.\ Claude's 24\%/16\%, Table~\ref{tab:action-dist}). The Claude-specific pattern is qualitative, not volumetric. Claude reaches for \texttt{bash} to read app state in place of the rubric-graded UI side-effect, the console-script shortcut detailed below.
\end{itemize}

\paragraph{Console-script shortcuts (Claude-specific).} A failure pattern unique to the two Claude models is the console-script shortcut. The agent opens a JavaScript console (or, given Anthropic's native \texttt{bash} tool, hits the app's REST endpoint directly with \texttt{curl}) and reads the persona's data without driving the visible UI. When the rubric only requires that the agent \emph{know} the value, this satisfies it. When the rubric requires a user-visible side-effect (drag a card, open the project, save a file from the menu), the script reads the data and \texttt{DONE}s the task without producing the artifact. \texttt{hard\_app-f026} (the 55-round \texttt{curl} trajectory above) is canonical. The judge notes that the agent ``investigates SprintBoard throughout the trajectory using API calls in the browser console\dots\ but it never actually opens the three SprintBoard projects.''

\paragraph{Skipped-app concrete examples.} On 323 failed rubrics, the agent finishes a multi-app task without ever opening one of the named apps. On \texttt{long\_horizon-f066} the agent archives a HooliChat conversation and never opens OddsMarket. On \texttt{long\_horizon-f074} it visits nine apps but never opens TableFind to make the required reservation (Figure~\ref{fig:vignettes-b}, row 04). On \texttt{aggregation-f010} it searches HooliChat extensively but never reads the Dundies categories file in \texttt{\char`~/Documents}.

\section{Persona Specification and Event Chains}
\label{app:persona}

The canonical \persona{} persona is stored as a single JSON document
(\texttt{personas/michael\_scott.json}, shipped in the release) with sixteen
top-level sections that the generator reads in dependency order. Every
seeded record in the \numapps{}-app environment can be traced back to one
of these sections.

\paragraph{Top-level schema.}\mbox{}\\[-2pt]
\begin{lstlisting}[style=promptstyle]
{
  "identity":           { name, age, city, address, employer, role,
                          salary, email, phone, gender, bio, ... },
  "contacts":           [ { name, relationship, frequency, email, phone,
                            shared_activities, apps_present_in,
                            message_personality, birthday, address }, ... ],
  "financial":          { checking_balance, savings_balance, credit_limit,
                          credit_used, monthly_income_net,
                          recurring_charges },
  "record_counts":      { per-app seeded row counts (generator output) },
  "investments":        { cash_balance, holdings, order_history, dividends },
  "prediction_markets": { balance, total_invested, net_pnl,
                          active_positions, watchlist },
  "routines":           { commute, exercise, meals, improv_class },
  "trips":              [ { destination, dates, hotel, flights, ... }, ... ],
  "work":               { projects: [ ... ] },
  "tax_info":           { tax_year, w2, freelance_1099, deductions,
                          state_code },
  "planted_contradictions": [ ... ],
  "planted_dependencies":   [ ... ],
  "browsing_patterns":      { research_threads, routine_browsing,
                              humor_searches },
  "shopping":               { online_orders, wishlist },
  "app_overrides":          { hoolishop, lockedin, batbucks, speedtax,
                              hoolichat, etaxi, hangrydash, tablefind,
                              ... },
  "cross_app_events":       [ ... ]
}
\end{lstlisting}

The four sections that drive cross-app consistency are
\texttt{cross\_app\_events} (cross-app side-effects, e.g.\ a trip seeds rows
in six apps), \texttt{planted\_contradictions} (deliberate red herrings that
test whether agents read all sources), \texttt{planted\_dependencies}
(records the rubric needs the agent to chain through), and
\texttt{app\_overrides} (per-app tuning, e.g.\ a Cheskepdia booking that a
later HangryDash record references).

\paragraph{Annotated event chain.} A single \texttt{cross\_app\_events}
entry produces correlated rows across every application that would
plausibly record the event. Below is the canonical Cooper's Seafood House
dinner-plan event, reproduced from the seed (internal app slugs translated
to their product names).

\begin{lstlisting}[style=promptstyle]
{
  "type": "dinner_plan",
  "description": "Romantic dinner at Cooper's Seafood House for Holly",
  "date": "2026-03-28",
  "time": "7:30pm",
  "apps": ["tablefind", "hoolichat", "gringotts", "hoolimail",
           "hoolicalendar"],
  "generates": {
    "hoolichat_mention": {
      "contact": "Jim Halpert",
      "context": "Jim. JIM. I need your help. I'm taking Holly to Cooper's
       tonight. What do I wear? Should I bring flowers? Is it too much if
       I also bring a boombox? Please respond immediately."
    },
    "browser_history": [
      "coopers seafood house scranton reviews",
      "romantic restaurants scranton",
      "how to be charming at dinner wikihow",
      "what wine goes with steak date night"
    ],
    "calendar_event": true
  }
}
\end{lstlisting}

\noindent The seeders fan out. The web-app seeder writes the
TableFind reservation and the Gringotts charge, the calendar seeder
writes the HooliCalendar block, the browser seeder writes the
Firefox history rows, and the chat seeder writes the message thread.
Because every seeder reads from the same event record, the entire chain
stays internally consistent.

\section{Task-Review Interface}
\label{app:viewer}

\begin{figure}[t]
  \centering
  \includegraphics[width=\textwidth]{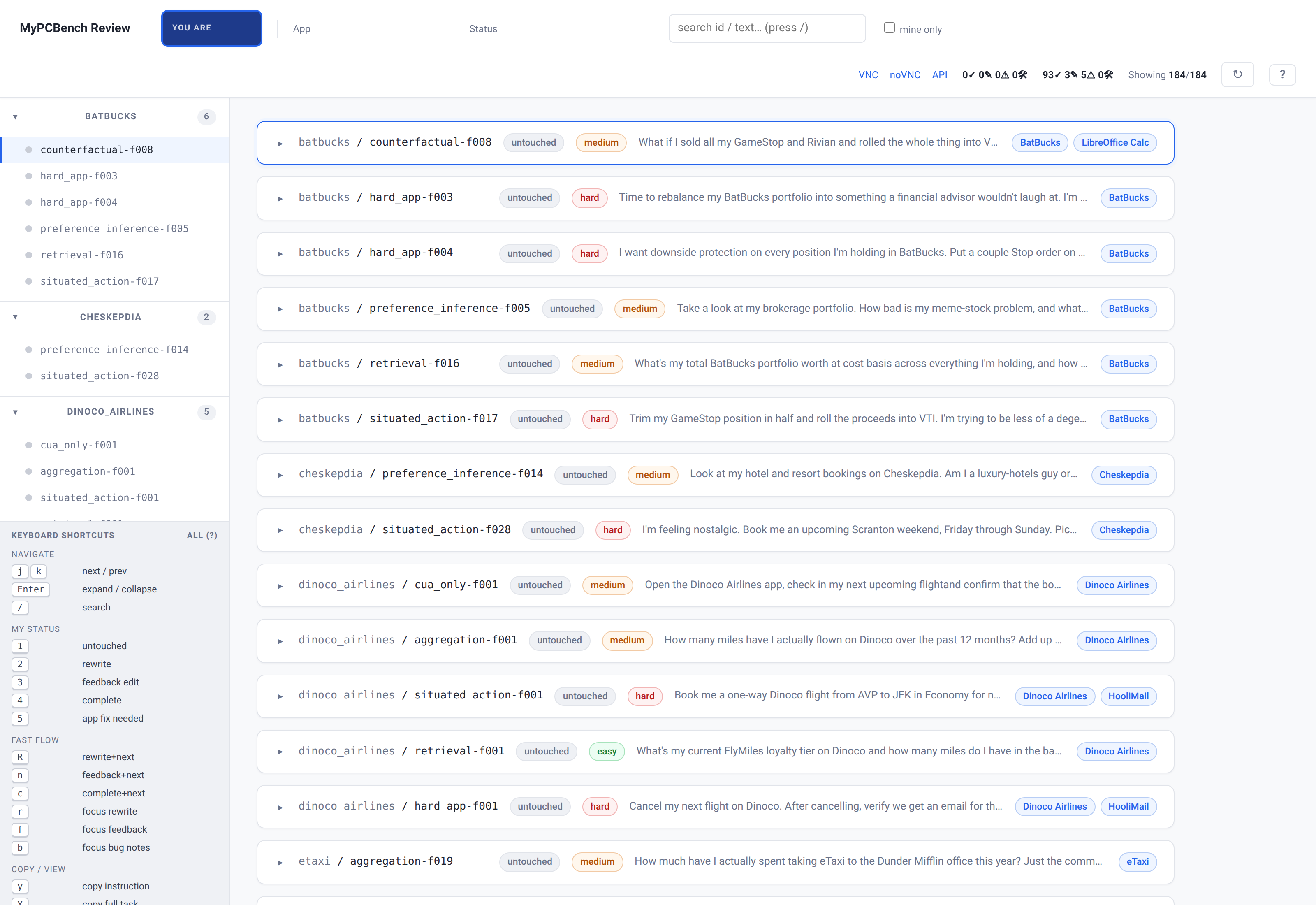}
  \caption{The \bench{} task-review interface used during quality assurance. The left pane is grouped by primary application. Each row shows the task identifier, review state, difficulty, the verbatim instruction preview, and pills for the apps the task touches.}
  \label{fig:reviewer-ui}
\end{figure}

We built a single-page web reviewer (Figure~\ref{fig:reviewer-ui}) for the quality-assurance pass described in \S\ref{sec:qa}. The interface lists every task in the suite grouped by primary application, surfaces the instruction, difficulty, and the apps the task touches inline, and exposes per-task review state with one-key keyboard shortcuts. Selecting a task expands a side-pane with the rubric items and a deep-link to the corresponding live application URL inside the VM, so a reviewer can run the task end-to-end without leaving the page.

\section{Data Generation Pipeline}
\label{app:datagen}

The pipeline turns the persona JSON in Appendix~\ref{app:persona} into a
fully populated Linux desktop image. A single deterministic entry point
calls one seeder per data surface in dependency order. Every seeder
consumes the same persona document, so adding a new persona is a
one-file change.

\paragraph{Seeders.} The seeders run in dependency order, each
consuming the same persona document. The core ones write each data
surface from the spec. The web-app seeder writes the SQLite databases
for the \numapps{} Next.js apps and honors \texttt{cross\_app\_events}
so a single trip leaves correlated rows across Cheskepdia, Dinoco,
eTaxi, Gringotts, HooliMail, and HooliCalendar. Others build the on-disk
Maildir at \texttt{/home/user/Maildir} and its HooliMail mirror, write
HooliCalendar events with the \texttt{.ics} files LibreOffice reads,
populate the Firefox profile and the per-app session cookies that keep
every app pre-logged-in, and place the persona's documents (meeting
notes, expense reports, itineraries, boarding passes, resumes) under
\texttt{/home/user}. Earlier steps resolve the active persona and
expand the JSON into derived fields such as paystub line items, and a
final step emits a Markdown persona summary used by reviewers and as
background context for the judge.

\paragraph{Determinism.} Every seeder is seeded from a deterministic RNG
keyed on the persona name and the seeder identifier. The reference time
defaults to the bake date (so the seeded data reads as current relative to
the build) and can be pinned. With the anchor pinned, identical inputs
produce byte-identical outputs across runs and machines.

\paragraph{From persona to image.} After all seeders run, a single
\texttt{docker build} bakes the populated home directory, the Firefox
profile, and the \numapps{} Next.js apps into the released environment
image. The first boot of the QEMU guest captures a base snapshot. Every
subsequent task starts from this snapshot, so the agent always sees the
same initial state.

\section{Grading and Rubric Prompts}
\label{app:prompts}

This appendix reproduces, verbatim from the released code, the prompts used to grade every task.

\paragraph{Judge system prompt.}\mbox{}\\[-2pt]
\begin{lstlisting}[style=promptstyle]
You are an expert evaluator of desktop-agent trajectories.

You will receive:
- The user task (for context).
- ONE specific rubric item with a criterion and (optional) verification description.
- The agent's full action history (one line per step).
- Every screenshot from the trajectory, in chronological order.

Your goal is to decide whether this single rubric item is satisfied by the trajectory.

Evaluation rules:
- Judge ONLY the one rubric item you are given; ignore all other implicit requirements.
- Ground your judgment in what the screenshots and actions actually show. Do not invent state.
- Filtering / sorting / form requirements must be applied AND confirmed (visible) to count as satisfied.
- If the agent was blocked (captcha, access denied, crash, etc.) and therefore could not satisfy the rubric, report failure.
- If a later step UNDID the rubric (e.g. user-visible state was correct, then was overwritten with wrong data), report failure.

Respond in exactly this format:

Thoughts: <your reasoning, citing specific steps/screenshots>
Status: "success" or "failure"
\end{lstlisting}

\paragraph{Judge user prompt (one call per rubric).} The user message is instantiated from the template below with the task instruction, the single rubric item being evaluated, and the agent's compacted action history. Up to the most recent 200 screenshots from the trajectory are attached in chronological order in the same message. When the rubric carries a \texttt{verification} note or a non-default \texttt{weight}, two additional lines (\texttt{Verification: ...} and \texttt{Weight: ...}) are emitted between \texttt{Requirement} and \texttt{Full Action History}.
\begin{lstlisting}[style=promptstyle]
User Task (context only): {task_instruction}

Evaluate ONLY this rubric item:
Rubric ID: {rubric_id}
Requirement: {rubric_criterion}

Full Action History:
{action_history}

Screenshots attached below: {n_screenshots} (trajectory had {n_steps} total step(s)).

Decide whether the rubric ({rubric_id}) is satisfied. Use the required 'Thoughts:' / 'Status:' format.
\end{lstlisting}

\paragraph{Aggregation.} Letting $s_r\in\{0,1\}$ denote whether the judge returned ``success'' for rubric $r$ and $w_r$ the authored rubric weights (normalized to sum to one within the task), the two reported metrics are
\[
\text{rubric score} = \sum_{r=1}^{N} w_r\,s_r, \qquad \text{perfect} = \mathbb{1}\!\left[\forall r:\, s_r = 1\right].
\]

\section{Agent Harness}
\label{app:harness}

This appendix documents the released agent harness, covering its interface, action space (Table~\ref{tab:actions}), step budgets, snapshot reset, and the actual system prompts used for each evaluated model family. Much of this is rehashed from the main section. 

\paragraph{Harness interface.} The harness spins up the Docker image (Ubuntu 24.04 with a GDM auto-login GNOME session, with the \numapps{} Next.js apps running as systemd services), waits for desktop-ready (typically ${\sim}$90~s), and enters the step loop. Each step fetches a screenshot through the guest's OSWorld-compatible HTTP Control API on port 5000 (\texttt{GET /screenshot}), constructs the agent message (system prompt + task instruction + screenshot + accumulated history), dispatches to the model's native API, parses the returned action, and executes it on the guest through the same Control API (\texttt{POST /execute}) or its shell endpoint. The loop repeats until the agent emits \texttt{done}/\texttt{DONE} or the step budget is reached. VNC/noVNC is exposed only for human observation of a running agent. A fresh base snapshot is restored between tasks (a copy-on-write overlay rebuild that matches OSWorld's \texttt{revert\_to\_snapshot}) so each task sees an identical initial state. The harness exposes two interchangeable backends. \texttt{--backend qemu} drives a QEMU/KVM guest directly from the host (default), while \texttt{--backend docker} runs the same guest inside the released Docker image for portability.

\paragraph{Action space.} Table~\ref{tab:actions} enumerates every action exposed to an evaluated agent. The top block is the unmodified OSWorld \texttt{pyautogui} surface and is mapped onto every provider's CUA vocabulary. The bottom block lists non-\texttt{pyautogui} tools. \texttt{bash} is exposed to every cua+bash agent, while Anthropic's file-edit tool remains Claude-only because the other provider APIs do not document an equivalent.

\begin{table}[h]
\centering
\caption{The \bench{} action space.}
\label{tab:actions}
\renewcommand{\arraystretch}{1.10}
\setlength{\tabcolsep}{4pt}
{\footnotesize
\begin{tabular}{@{}p{0.30\textwidth}p{0.36\textwidth}p{0.24\textwidth}@{}}
\toprule
\textbf{Action} & \textbf{Parameters} & \textbf{Available to} \\
\midrule
\texttt{click}, \texttt{double\_click}, \texttt{right\_click} & $x, y$ pixel coordinates & all CUA agents \\
\texttt{type} & text string & all CUA agents \\
\texttt{key} & key combination (e.g., \texttt{ctrl+c}) & all CUA agents \\
\texttt{scroll} & $x, y$, scroll amount & all CUA agents \\
\texttt{drag} & start $(x, y)$, end $(x, y)$ & all CUA agents \\
\texttt{wait} & duration (seconds) & all CUA agents \\
\texttt{screenshot} & --- & all CUA agents \\
\texttt{done} / \texttt{fail} & --- & all CUA agents \\
\midrule
\texttt{bash} & shell command string & all cua+bash agents \\
\texttt{str\_replace\_based\_edit\_tool} & view / create / replace / insert & Claude (native) \\
\bottomrule
\end{tabular}
}
\end{table}

\begin{figure}[h]
\centering
\includegraphics[width=\textwidth]{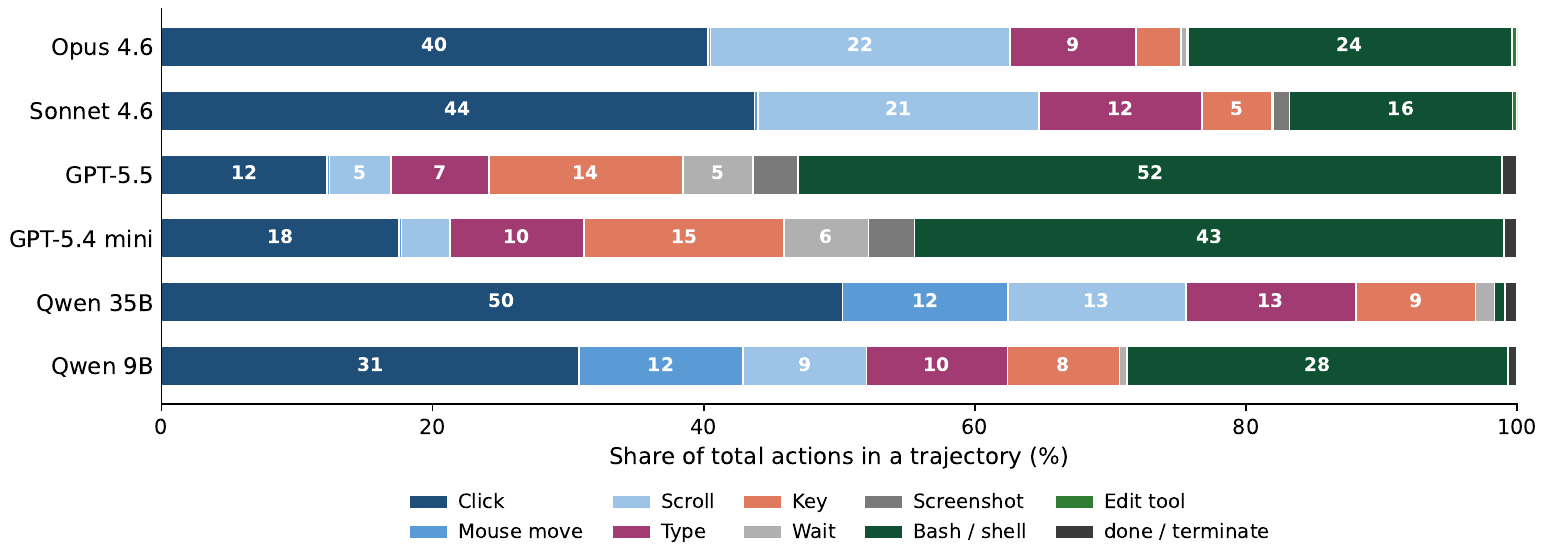}
\caption{Per-model action distribution across all \numtasks{} trajectories, grouped into the categories listed in the legend. All six models share the same \texttt{computer}+\texttt{bash} surface, but the families use it very differently.}
\label{fig:action-dist}
\end{figure}

\paragraph{Action distribution by model.}\label{para:action-dist} For each of the six evaluated models we recover the action sequence emitted on every one of the \numtasks{} tasks. Claude actions come from the \texttt{tool\_use} blocks in each \texttt{messages.json}. For GPT and Qwen we take one record per executed action from each \texttt{traj.jsonl}, selecting per task the canonical run whose record count matches the published step count (the table caption notes the handful of closest-match exceptions). \texttt{pyautogui} dispatch is reverse-mapped to the provider's \texttt{computer.*} action. We collapse \texttt{moveTo+click} pairs into a single click so click counts line up across surfaces (the OSWorld-style Qwen surface emits \texttt{mouse\_move} as a separate action by construction), and every non-\texttt{pyautogui} tool-call round counts as a shell (\texttt{bash}) call, the only non-\texttt{computer} tool exposed in cua+bash mode. Figure~\ref{fig:action-dist} visualizes the resulting distribution, and Table~\ref{tab:action-dist} gives the per-action shares.

\begin{table}[h]
\centering
\caption{Full per-model action distribution across all \numtasks{} trajectories. Each cell is the share of that model's total actions, and the bottom rows give trajectory counts and absolute totals. For GPT and Qwen, \texttt{bash} counts every non-\texttt{pyautogui} tool-call round (the only non-\texttt{computer} tool in cua+bash mode), so the figure measures shell \emph{reliance}, not per-command granularity. \texttt{str\_replace\_based\_edit\_tool} is Claude-only, as the other providers ship no documented equivalent. \texttt{triple\_click} only appears in the Claude action surface. The OSWorld-style Qwen pipeline emits \texttt{mouse\_move} as a separate action (Claude's and OpenAI's APIs fold movement into the click), inflating Qwen's mouse-only share to $\sim$12\%. The canonical trajectory per task is the single run whose record count matches the published step count --- no best-of-N. All models cover 184/184 tasks. For GPT-5.4 mini, Qwen 35B, and Qwen 9B, 3 / 21 / 26 tasks had no run whose record count exactly matched and the closest was used.}
\label{tab:action-dist}
\renewcommand{\arraystretch}{1.05}
\setlength{\tabcolsep}{6pt}
{\footnotesize
\begin{tabular}{@{}lrrrrrr@{}}
\toprule
\textbf{Action} & \textbf{Opus 4.6} & \textbf{Sonnet 4.6} & \textbf{GPT-5.5} & \textbf{GPT-5.4 mini} & \textbf{Qwen 35B} & \textbf{Qwen 9B} \\
\midrule
\texttt{left\_click}        & 37.6\% & 41.7\% & 11.8\% & 16.9\% & \textbf{48.1\%} & \textbf{29.5\%} \\
\texttt{double\_click}      & \phantom{0}0.5\% & \phantom{0}0.1\% & \phantom{0}0.2\% & \phantom{0}0.6\% & \phantom{0}1.1\% & \phantom{0}1.3\% \\
\texttt{triple\_click}      & \phantom{0}2.0\% & \phantom{0}1.9\% & \phantom{0}0.0\% & \phantom{0}0.0\% & \phantom{0}0.0\% & \phantom{0}0.0\% \\
\texttt{right\_click}       & \phantom{0}0.1\% & \phantom{0}0.0\% & \phantom{0}0.0\% & \phantom{0}0.0\% & \phantom{0}0.4\% & \phantom{0}0.1\% \\
\texttt{drag}               & \phantom{0}0.0\% & \phantom{0}0.1\% & \phantom{0}0.2\% & \phantom{0}0.1\% & \phantom{0}0.7\% & \phantom{0}0.0\% \\
\texttt{mouse\_move}        & \phantom{0}0.2\% & \phantom{0}0.2\% & \phantom{0}0.2\% & \phantom{0}0.2\% & 12.2\% & 12.1\% \\
\midrule
\texttt{scroll}             & 22.1\% & 20.7\% & \phantom{0}4.5\% & \phantom{0}3.6\% & 13.1\% & \phantom{0}9.1\% \\
\texttt{type}               & \phantom{0}9.3\% & 12.0\% & \phantom{0}7.2\% & \phantom{0}9.9\% & 12.6\% & 10.4\% \\
\texttt{key}                & \phantom{0}3.3\% & \phantom{0}5.2\% & 14.3\% & 14.8\% & \phantom{0}8.8\% & \phantom{0}8.3\% \\
\texttt{wait}               & \phantom{0}0.4\% & \phantom{0}0.0\% & \phantom{0}5.1\% & \phantom{0}6.2\% & \phantom{0}1.4\% & \phantom{0}0.6\% \\
\texttt{screenshot}         & \phantom{0}0.1\% & \phantom{0}1.2\% & \phantom{0}3.4\% & \phantom{0}3.4\% & \phantom{0}0.0\% & \phantom{0}0.0\% \\
\texttt{done} / \texttt{terminate} & \phantom{0}0.0\% & \phantom{0}0.0\% & \phantom{0}1.1\% & \phantom{0}1.0\% & \phantom{0}0.9\% & \phantom{0}0.7\% \\
\midrule
\texttt{bash}               & 23.9\% & 16.4\% & \textbf{51.9\%} & \textbf{43.5\%} & \phantom{0}0.8\% & 28.1\% \\
\texttt{str\_replace\_based\_edit\_tool} & \phantom{0}0.4\% & \phantom{0}0.3\% & \phantom{0}0.0\% & \phantom{0}0.0\% & \phantom{0}0.0\% & \phantom{0}0.0\% \\
\midrule
Trajectories analyzed       & 184 & 184 & 184 & 184 & 184 & 184 \\
Total actions               & 8{,}575 & 8{,}584 & 8{,}420 & 8{,}148 & 12{,}587 & 13{,}623 \\
\bottomrule
\end{tabular}
}
\end{table}

\paragraph{What the action shapes say.} Figure~\ref{fig:action-dist} shows how the families use the \emph{same} \texttt{computer}+\texttt{bash} surface very differently when pointed at the same \numtasks{} tasks. \emph{Claude} treats the desktop as a stable hybrid. Around 70\% of actions go through the UI (click, scroll, type) and roughly 24\% through \texttt{bash} on Opus, sliding toward more click and less bash on Sonnet. That bash share is what \S\ref{sec:failures} catalogs as the console-script shortcut, reading app state via \texttt{curl} without moving the visible UI. \emph{GPT-5.5 and GPT-5.4 mini} are, despite the dual-tool hint, the \emph{most} shell-dependent family --- 52\% and 44\% of their actions are shell (\texttt{bash}) rounds, about double Claude's share, with a correspondingly small UI footprint (12--17\% click, 4--5\% scroll, 14--15\% \texttt{key}). \emph{The Qwen models split.} Qwen 35B essentially ignores the shell tool (0.8\% \texttt{bash}) and stays click-and-\texttt{mouse\_move} heavy (48\%+12\%), reflecting the OSWorld-style coordinate-emission surface it was trained against. Its failures are the hallucination and surface-error modes it leads (\S\ref{sec:failures}), not a tool-access artifact. Qwen 9B's 28\% \texttt{bash} is not productive shell use but the malformed dual-schema tool calls behind the zero-score collapse cataloged in Appendix~\ref{app:failures-detail}.

\paragraph{When \texttt{bash} helps and when it hurts.} Bash use is widespread in the Claude tier, with 155 of 184 Opus trajectories and 152 of 184 Sonnet trajectories invoking it at least once. It is not, on the whole, a positive predictor of task success. Among Opus trajectories, the perfect rate is 69.0\% (20/29) when \texttt{bash} is never invoked and 52.9\% (82/155) when it is. The corresponding Sonnet numbers are 43.8\% (14/32) versus 38.2\% (58/152). The gap is correlational rather than causal, because Claude reaches for \texttt{bash} more often on the harder, multi-app tasks where the perfect rate is lower to begin with. The comparison does illustrate the failure mode cataloged in \S\ref{sec:failures}. Bash is fast and reliable for \emph{reading} app state via REST endpoints, and it is the wrong tool when the rubric requires a user-visible side-effect such as moving a card or saving a file from the menu --- the 55-round \texttt{curl}-only trajectory on \texttt{hard\_app-f026} (Appendix~\ref{app:failures-detail}) is a clean instance.

\paragraph{Implications for hybrid CUA designs.} The action distributions (Figure~\ref{fig:action-dist}) show the same tool plus the same hint produces opposite behaviors across families. A hybrid agent therefore needs a policy for choosing between \texttt{bash} and the UI as a function of \emph{what the task needs to leave behind}, not just what answer it needs to produce, since rubrics that grade on user-visible side-effects can punish a correct \texttt{bash}-only path. The tool alone is not enough. Harness or prompt design must make explicit when UI grounding is required rather than leaving the choice fully to the agent. Shell access does \emph{not} generalize uniformly across CUA families, so a useful follow-up is per-family guidance (or harness-side constraints on side-effect-graded rubrics) rather than a single shared hint.

\paragraph{Shared persona-and-environment context (all agents).} Every evaluated agent, regardless of model family, receives the following persona-and-environment block appended to its system prompt. This is the shared frame that keeps the agents grounded on the correct persona, applications, and conventions. It is reproduced verbatim from the released code.
\begin{lstlisting}[style=promptstyle]
## Persona

- Name: Michael Scott
- Email: `michael.scott@dundermifflin.com`
- Linux user: `user` (sudo password: `{CLIENT_PASSWORD}`)

## Environment

- Ubuntu 24.04 GNOME desktop. Browser: Firefox (pre-logged-in to every
  web app via the bookmarks toolbar).
- Pinned to dock: HooliChat, HooliWork, Firefox, LibreOffice Writer/Calc/Impress, VS Code.
- `/home/user/Documents/`, `/home/user/Downloads/`, `/home/user/Maildir/` hold persona files.
- Python 3.12 and the LibreOffice CLI are available in the VM.

## Web apps

Each is served at `http://localhost:PORT`, pre-authenticated as the persona.

| Port | App             | Domain                                              |
|------|-----------------|-----------------------------------------------------|
| 3001 | Gringotts       | personal banking: accounts, transactions, transfers |
| 3002 | BatBucks        | stock / crypto trading: portfolio, orders           |
| 3003 | OddsMarket      | prediction markets: bets, positions                 |
| 3004 | HooliChat       | direct + group messaging                            |
| 3005 | HooliWork       | workplace channels                                  |
| 3006 | eTaxi           | ride hailing: trips, drivers                        |
| 3007 | HangryDash      | food delivery: orders, restaurants                  |
| 3008 | TableFind       | restaurant reservations                             |
| 3009 | Kwik-E-Mart     | grocery orders, inventory                           |
| 3010 | HooliShop       | e-commerce: orders, carts, products                 |
| 3011 | Dinoco Airlines | flight bookings, itineraries                        |
| 3012 | Cheskepdia      | short-term rental bookings                          |
| 3013 | SprintBoard     | project tasks, sprints                              |
| 3014 | LockedIn        | professional networking, jobs, connections          |
| 3015 | SpeedTax        | tax returns, filings                                |
| 3016 | HooliMail       | email inbox                                         |
| 3017 | HooliCalendar   | events, invitations                                 |

## Output

- Place your final answer (numbers, text, file paths) as plain text in
  your last assistant turn before any stop signal.
\end{lstlisting}

\paragraph{Task-completion discipline (all agents).} A single shared block on \emph{when not} to terminate, also appended to every system prompt (\texttt{COMPLETION\_DISCIPLINE} in \texttt{agents/prompts.py}).
\begin{lstlisting}[style=promptstyle]
Task completion discipline:
- Do NOT emit `DONE`, `terminate`, or any stop signal until you have actually completed the task. A task is only complete when you have produced the specific output the user asked for AND verified it looks correct.
- Use all available steps --- plan, act, observe, iterate. Don't bail out early just because the first approach didn't work.
- If something fails, try a different approach (different coordinates, different app, different bash command). Never give up on the first error.
- Always write your final answer (numbers, text, file contents) before terminating --- the grader reads your last response to check correctness.
- Only emit `FAIL` if the task is genuinely impossible (required data literally does not exist). Never use `FAIL` as a shortcut when the task is just hard.
\end{lstlisting}

\paragraph{Claude Computer Use system prompt.} Used by Claude Opus 4.6 and Claude Sonnet 4.6. The released prompt (\texttt{CLAUDE\_CUA\_SYSTEM\_PROMPT} in \texttt{agents/prompts.py}) follows Anthropic's recommended XML-tagged scaffold (\texttt{<role>} / \texttt{<instructions>} / \texttt{<safety>} / \texttt{<environment>}) and splices the shared \texttt{VERIFICATION\_GUIDANCE}, \texttt{PARALLEL\_TOOL\_HINT}, \texttt{SHELL\_NON\_INTERACTIVE\_NOTE}, and \texttt{KEYBOARD\_SHORTCUT\_HINT} blocks into \texttt{<instructions>}. The persona block above goes inside \texttt{<environment>}. The leading \texttt{<role>}/\texttt{<instructions>} text is reproduced below.
\begin{lstlisting}[style=promptstyle]
<role>
You are an AI agent operating a Linux workstation. Your tools are `computer` (screenshot + mouse/keyboard), `bash` (shell commands in the VM), and `str_replace_based_edit_tool` (file view/create/str_replace/insert).
</role>

<instructions>
Stop signal: state your final answer in plain text, then emit ```DONE``` (or ```FAIL``` / `[INFEASIBLE]` if impossible).
...
</instructions>
\end{lstlisting}

\paragraph{OpenAI CUA operator prompt.} Injected as the text portion of the first user message for OpenAI computer-use agents (GPT-5.5, GPT-5.4 mini). The released constant (\texttt{OPENAI\_CUA\_OPERATOR\_PROMPT} in \texttt{agents/prompts.py}) concatenates the lead-in below with \texttt{SAFETY\_PREAMBLE}, \texttt{VERIFICATION\_GUIDANCE}, \texttt{PARALLEL\_TOOL\_HINT}, \texttt{SHELL\_NON\_INTERACTIVE\_NOTE}, and \texttt{KEYBOARD\_SHORTCUT\_HINT}. The shared persona context is then appended by the caller.
\begin{lstlisting}[style=promptstyle]
You are an agent on a Linux desktop. Your tools are `computer` and `bash`.

Stop signal: state your final answer in plain text, then emit ```DONE``` (or ```FAIL``` / `[INFEASIBLE]` if the task is impossible).
\end{lstlisting}

\paragraph{Qwen tool-call system prompt.} Used for the open-weight agents (Qwen 3.5 35B-A3B and Qwen 3.5 9B). The vendored \texttt{qwen35vl\_agent} from OSWorld~\citep{xie2024osworld} emits a structured XML \texttt{<tool\_call>} per turn, and the harness parses each call into the corresponding OSWorld \texttt{pyautogui} dispatch. The system prompt below is reproduced verbatim from \texttt{agents/vendored\_paper\_results/qwen35vl\_agent.py}. The shared persona context (and, in cua+bash mode, an additional \texttt{bash}-tool description) is appended on the same system message.
\begin{lstlisting}[style=promptstyle]
You are a multi-purpose intelligent assistant. Based on my requests, you can use tools to help me complete various tasks.

# Tools

You have access to the following functions:

<tools>{tools_json}</tools>

If you choose to call a function ONLY reply in the following format with NO suffix:

<tool_call>
<function=example_function_name>
<parameter=example_parameter_1>
value_1
</parameter>
<parameter=example_parameter_2>
This is the value for the second parameter
that can span
multiple lines
</parameter>
</function>
</tool_call>

<IMPORTANT>
Reminder:
- Function calls MUST follow the specified format: an inner <function=...></function> block must be nested within <tool_call></tool_call> XML tags
- Required parameters MUST be specified
- You may provide optional reasoning for your function call in natural language BEFORE the function call, but NOT after
- If there is no function call available, answer the question like normal with your current knowledge and do not tell the user about function calls
- The current date is {today}.
- Collapsed screenshots appear as text: {collapse_text}
</IMPORTANT>

# Response format

Response format for every step:
1) Action: a short imperative describing what to do in the UI.
2) A single <tool_call>...</tool_call> block.

Rules:
- Output exactly in the order: Action, <tool_call>.
- Be brief: one sentence for Action.
- Do not output anything else outside those parts.
- If finishing successfully, use action=terminate with status=success in the tool call.
- If the task is infeasible or impossible, use action=terminate with status=failure in the tool call.
\end{lstlisting}

\section{Example Trajectories}
\label{app:trajectories}

\begin{figure}[p]
\centering
\includegraphics[width=\textwidth,height=0.95\textheight,keepaspectratio]{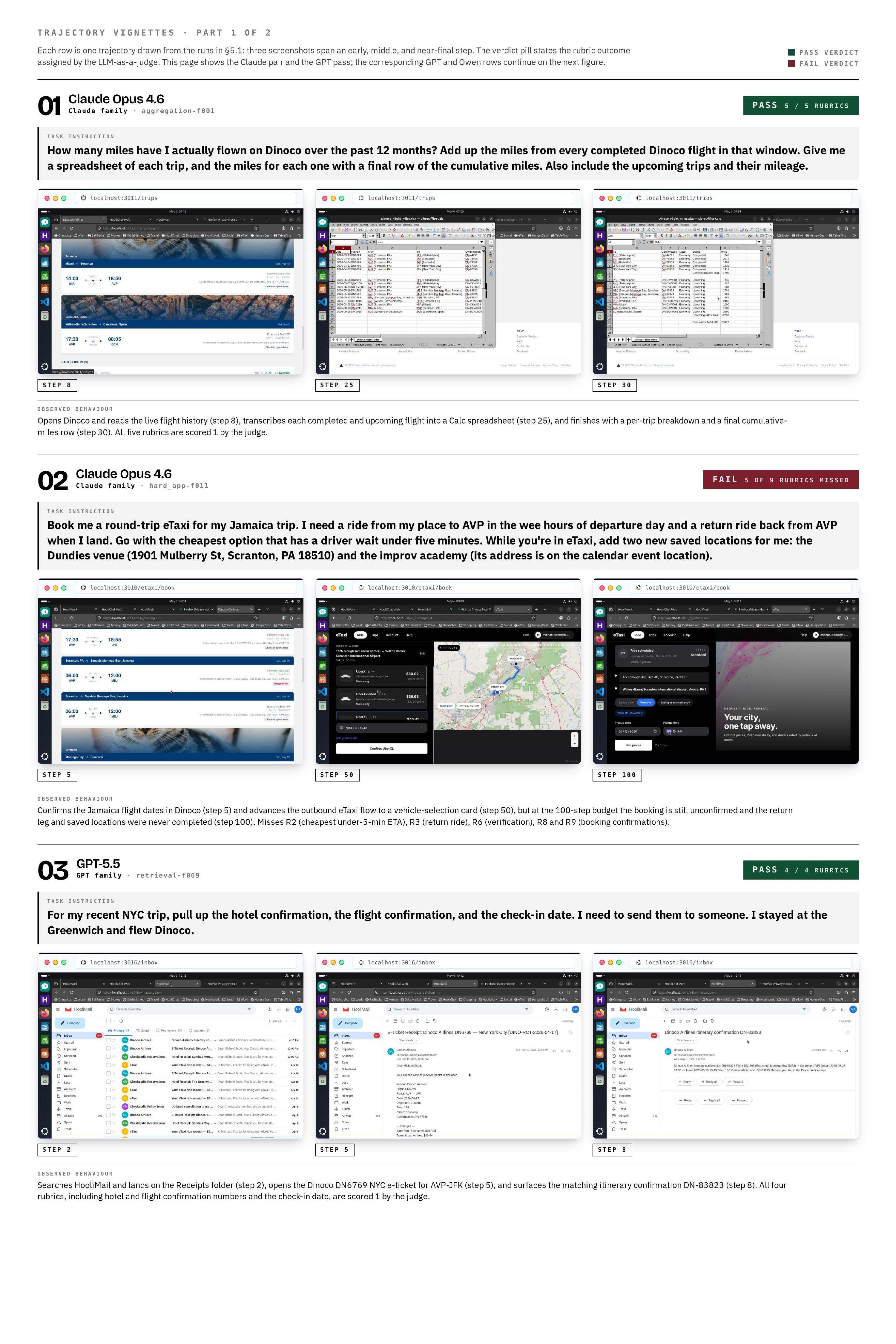}
\caption{Trajectory vignettes, part 1 of 2. Rows: Claude Opus 4.6 PASS on \texttt{aggregation-f001}, Claude Opus 4.6 FAIL on \texttt{hard\_app-f011}, GPT-5.5 PASS on \texttt{retrieval-f009}. Each row shows the verbatim task instruction, three screenshots (early, mid, near-final step), and an observed-behavior note linked to the rubric outcome assigned by the LLM-as-a-judge. The GPT row is taken from the computer-only (cua-only) configuration runs of the appendix ablation, chosen as representative of the family behaviors discussed in the main results.}
\label{fig:vignettes-a}
\end{figure}

\begin{figure}[p]
\centering
\includegraphics[width=\textwidth,height=0.95\textheight,keepaspectratio]{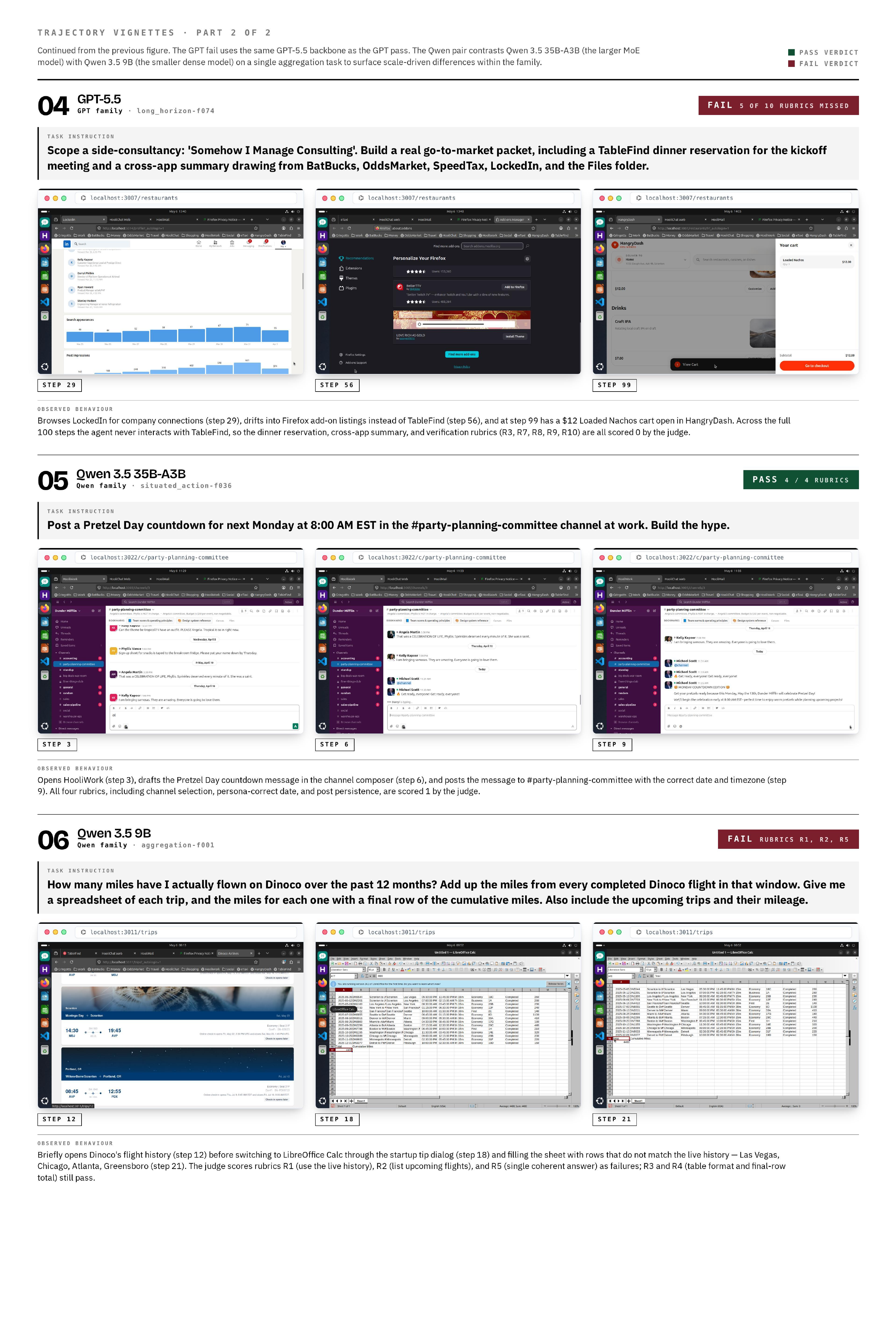}
\caption{Trajectory vignettes, part 2 of 2 (continued from Figure~\ref{fig:vignettes-a}). Rows: GPT-5.5 FAIL on \texttt{long\_horizon-f074}, Qwen 3.5 35B-A3B PASS on \texttt{situated\_action-f036}, Qwen 3.5 9B FAIL on \texttt{aggregation-f001}. Same layout convention as Figure~\ref{fig:vignettes-a}. The GPT and Qwen rows are taken from the computer-only (cua-only) configuration runs of the appendix ablation, chosen as representative of the family behaviors discussed in the main results.}
\label{fig:vignettes-b}
\end{figure}

Figures~\ref{fig:vignettes-a} and~\ref{fig:vignettes-b} pair one passing and one failing trajectory from each of the three evaluated model families (Claude, GPT, Qwen) at higher fidelity than the main-paper Figure~\ref{fig:walkthrough}. Each vignette row reproduces the verbatim task instruction issued to the agent, three screenshots from the actual trajectory (one early, one mid, one near the end), and a short observed-behavior note explaining how the trajectory arrives at the judge verdict shown in the pill. The six selected runs cover \texttt{aggregation-f001}, \texttt{hard\_app-f011}, \texttt{retrieval-f009}, \texttt{long\_horizon-f074}, and \texttt{situated\_action-f036}, drawn from three of the failure modes and the family-level trends discussed in \S\ref{sec:failures}. The Qwen pair places the larger MoE model (Qwen 3.5 35B-A3B) and the smaller dense model (Qwen 3.5 9B) on the same aggregation task to surface scale-driven differences within the family.

\clearpage

\section{Limitations}
\label{app:limitations}

\bench{} commits to one canonical persona (\persona{}) and one Linux/GNOME/Firefox software stack. The benchmark chooses depth over persona diversity. It measures whether agents can use one coherent personal computer deeply, not whether performance generalizes across demographics, locales, or device stacks. Grading uses a single Gemini judge. Absolute failure-mode counts (Table~\ref{tab:failure-modes}) should therefore be read as a structural breakdown across the six evaluated models, not as a precise prevalence estimate. The seeded persona is intentionally low-sensitivity (a public fictional character), so behaviors that emerge only on agents reasoning over genuinely sensitive personal data are not exercised by this benchmark. We treat that as an explicit out-of-scope choice.

\section{Broader Impact}
\label{app:broader-impact}

A personally intelligent agent is, by construction, an agent that can act on a user's full digital life. On the upside, a benchmark that explicitly tests cross-app, cross-history personalization should make it harder to ship assistants that look competent on stock-state demos but fail the moment they meet real personal data, and the failure-mode catalog identifies concrete behaviors for developers to measure and reduce (premature \texttt{DONE}, surface-error abandonment, skipped apps, hallucinated persona values, console-script shortcuts). On the downside, the same skills that drive a clean Dundies-lifecycle plan against Michael Scott's seeded desktop are the skills required to drive an agent against a real user's logged-in accounts. Numbers on this benchmark should not be read as a clearance to deploy CUA agents on production accounts. We mitigate the immediate dual-use surface by (i)~seeding only synthetic data tied to a public fictional persona, so the released image contains no real PII or real correspondence, (ii)~hosting every web application locally inside the QEMU guest, so credentials and form values cannot be exfiltrated to the live web during evaluation, and (iii)~recommending offline benchmarking against the released image as the intended use, and only that. The eval VM does retain outbound network access, and an optional host-provided \texttt{OPENAI\_API\_KEY} can be injected to power in-character NPC chat replies inside the messaging apps (the feature is disabled when no key is provided).

\section{Release Artifacts}
\label{app:release}

The project page (\url{https://mypcbench.com}) and code repository (\url{https://github.com/ljang0/MyPCBench}) host (i)~the environment image (Docker + QEMU snapshot), (ii)~the set of \numtasks{} task evaluations, (iii)~the per-task rubrics, (iv)~the agent harness that connects standard CUA agents to the environment, (v)~the configuration of the rubric-grading judge, and (vi)~the persona specification (\texttt{personas/michael\_scott.json}). We do \emph{not} release the agent trajectories or per-rubric judge outputs produced by the runs in this paper. Any future work using the same harness and judge can reproduce them on the released image.

\end{document}